\documentclass{article}

\PassOptionsToPackage{numbers, compress}{natbib}

\usepackage[preprint]{neurips_2025}

\usepackage[utf8]{inputenc} 
\usepackage[T1]{fontenc}    
\usepackage{hyperref}       
\usepackage{url}            
\usepackage{booktabs}       
\usepackage{amsfonts}       
\usepackage{nicefrac}       
\usepackage{microtype}      
\usepackage{xcolor} 
\usepackage{wrapfig}
\usepackage{graphicx}
\usepackage{amsmath}
\usepackage{amsthm}
\usepackage{threeparttable}
\usepackage{longtable}
\usepackage{supertabular}
\usepackage{multirow,multicol}
\usepackage{subcaption}
\usepackage{listings}
\newtheorem{proposition}{Proposition}

\newtheorem{definition}{Definition}
\newtheorem{observation}{Observation}
\theoremstyle{remark}
\newtheorem*{remark}{Remark}
\usepackage{bm}
\usepackage{comment}

\title{Towards Principled Task Grouping \\for Multi-Task Learning
}

\author{%
  Chenguang Wang\textsuperscript{1,†}, 
  Xuanhao Pan\textsuperscript{1,†}, 
  Tianshu Yu\textsuperscript{1} \\
  \texttt{\{chenguangwang,xuanhaopan\}@link.cuhk.edu.cn, yutianshu@cuhk.edu.cn} \\
  \textsuperscript{1}School of Data Science, The Chinese University of Hong Kong, Shenzhen\\
  \textsuperscript{†}\footnotesize{Equal Contribution.}
}

\begin{document}

\maketitle

\begin{abstract}
Multi-task learning (MTL) aims to leverage shared information among tasks to improve learning efficiency and accuracy. However, MTL often struggles to effectively manage positive and negative transfer between tasks, which can hinder performance improvements. Task grouping addresses this challenge by organizing tasks into meaningful clusters, maximizing beneficial transfer while minimizing detrimental interactions. 
This paper introduces a principled approach to task grouping in MTL, advancing beyond existing methods by addressing key theoretical and practical limitations. Unlike prior studies, our method offers a theoretically grounded approach that does not depend on restrictive assumptions for constructing transfer gains. We also present a flexible mathematical programming formulation that accommodates a wide range of resource constraints, thereby enhancing its versatility.
Experimental results across diverse domains, including computer vision datasets, combinatorial optimization benchmarks, and time series tasks, demonstrate the superiority of our method over extensive baselines, thereby validating its effectiveness and general applicability in MTL without sacrificing efficiency. 
\end{abstract}
\section{Introduction}\label{sec: intro}
Multitask Learning (MTL) \cite{caruana1997multitask,zhang2021survey,vandenhende2021multi} represents a forefront area in machine learning, aiming to improve learning efficiency and prediction accuracy by leveraging commonalities and differences across multiple tasks, reflected by the so-called intra-task ``transfer gain''. 
Building upon this foundational concept, MTL has exhibited exceptional performance across a spectrum of domains, including computer vision \cite{standley2020tasks, fifty2021efficiently,song2022efficient,sherif2023stg}, NLP \cite{zhang2022survey,Ding2023MitigatingTI}, Neural Architecture Search \cite{guo2020learning,Zhang2022ATM,Raychaudhuri2022ControllableDM,Yue2023LearningCA},  speech recognition \cite{Zhang2019AttentionaugmentedEM,huang2022mtl} and combinatorial optimization problems \cite{wang2023efficient,wang2024asp}.
Central to the optimization of this framework is the concept of task grouping. Task grouping \cite{kang2011learning,kumar2012learning,lee2016asymmetric, lee2018deep, zamir2018taskonomy, dwivedi2019representation,malhotra2022dropped,standley2020tasks,fifty2021efficiently, song2022efficient} in MTL involves strategically dividing a set of tasks into several groups, where each group encapsulates tasks that share maximal positive transfer while minimizing negative transfer. 

Recent studies \cite{standley2020tasks, fifty2021efficiently}  have contributed significantly to this domain. Both works utilize a methodology where specific task affinities are collected in a single run of the MTL training, which is then used to group tasks based on the assumption of high-order approximation on the task relationships. Subsequently, these groups are trained separately using MTL methods. However, these approaches exhibit key limitations. Firstly, there is an absence of a theoretical guarantee in their task affinity measures, raising concerns about the reliability and predictability of the task grouping effectiveness. Secondly, they rely on an enumeration-based branch and bound algorithm for solving the task grouping problem. This approach not only sacrifices efficiency in terms of computational resources but also poses challenges in incorporating additional constraints, limiting its practical applicability in more complex and realistic scenarios.

In this work, we introduce a novel approach to task grouping in MTL that addresses existing limitations and offers significant advancements over current methodologies.  {First}, we propose a theoretically grounded method for constructing transfer gains. Unlike TAG \cite{fifty2021efficiently}, which assumes restrictive conditions such as convexity and smoothness on loss functions, the proposed transfer gain is derived independently of any conditions.
Additionally, it maintains computational complexity at the same order as TAG by adhering to the high-order approximation assumption regarding task relationships, as utilized in prior work \cite{standley2020tasks,fifty2021efficiently}.
{Second}, our work introduces a  generic and flexible mathematical programming formulation to solve task grouping problems. This formulation can readily incorporate various budget constraints, a critical aspect of real-world applications.
By doing so, our method ensures the practicality and adaptability of MTL models in diverse scenarios, ranging from computational budget allocation to resource utilization considerations.

Our experimental evaluations across various domains, including computer vision datasets, combinatorial optimization benchmarks, and time series datasets, demonstrate the validity and generality of our proposed task grouping strategy in three key aspects.
First, we establish that our method consistently outperforms a wide range of baselines, encompassing single-task learning, multi-task learning, and various task grouping methods. This substantiates its effectiveness across these three diverse domains. 
Second, we illustrate the flexibility and effectiveness of our proposed mathematical programming formulation by introducing various constraints, mirroring real-world scenarios where resource budgets, such as GPU memory limitations and resource utilization, come into play. 
Our results demonstrate that our method significantly outperforms the baseline methods, showcasing its adaptability and performance improvement under such constraints. 
Finally, we propose two efficiency enhancing strategies, a sampling approach and a lazy collection mechanism, that substantially reduce the computational overhead of task grouping while maintaining performance quality.

In summary, this work makes several key contributions to the realm of task grouping: \textbf{({1})} We propose a theoretically principled method for constructing transfer gains without relying on restrictive assumptions; \textbf{({2})} We introduce a generic and flexible mathematical programming formulation capable of seamlessly integrating various budget constraints to solve task grouping problems; \textbf{({3})} Through extensive experiments, we demonstrate the effectiveness of our task grouping strategy across various domains and empirically showcase the flexibility of our mathematical programming approach by addressing realistic constraints.


\section{Related Works}\label{sec: related works}

\textbf{Task Grouping.} The idea of task grouping is to exploit shared knowledge within each group to improve overall learning efficiency. 
Early works utilized normalization terms to partition model parameters aligned with task groups \cite{kang2011learning,kumar2012learning,lee2016asymmetric}.
\cite{lee2018deep} extended this approach to deep learning, modeling asymmetric task relationships via autoencoders.
\cite{zamir2018taskonomy} presented ``Taskonomy'', which disentangles task relationships based on transfer learning hierarchies.
\cite{dwivedi2019representation} introduced representation similarity analysis for task taxonomy, demonstrating effectiveness on the Taskonomy dataset. 
\cite{malhotra2022dropped} introduced scheduled task mitigation to dynamically sequence task learning.
Closely related to our work, \cite{standley2020tasks,fifty2021efficiently} apply a two-stage methodology: first collecting training information and defining task affinities, then using Branch and Bound algorithms to find optimal groupings. More recent approaches include meta-learning for estimating grouping gains \cite{song2022efficient}, data-driven methods based on Data Maps \cite{sherif2023stg}, differentiable task grouping (DMTG) \cite{gao2024dmtg}, and gradient-based methods for affinity estimation (Grad-TAG) \cite{li2024Scalable}.

\textbf{Lookahead Methods.} The philosophy of Lookahead methods is using future information to guide the current state, which has been widely used in meta-learning \cite{finn2017model, nichol2018first,wang2020negative}, multitask learning \cite{fifty2021efficiently} and optimization techniques \cite{zhang2019lookahead,wang2020lookahead,zhou2021towards,byun2022multi}. In particular for multitask learning, \cite{fifty2021efficiently} collected the one-step-forward loss information between task pairs for each gradient updating and constructed the overall task affinity matrix at the end of training. We follow this idea but construct a more principled lookahead-based metric.

\textbf{Loss Balance.} Numerous works have emerged to address multitask learning by exploring the balance on the losses from different tasks \cite{mao2021banditmtl,yu2020gradient,DBLP:conf/iclr/JavaloyV22,navon2022multi,kendall2018multi,liu2021conflict,liu2021towards,guangyuan2022recon,DBLP:journals/tmlr/LiuJDJ22}. In these works, various loss reweighing mechanisms are designed to dynamically balance the importance of each task, which can relieve the negative transfer among tasks in terms of gradient information. Being plug-and-play techniques, loss balance methods can be flexibly applied to existing multitask learning frameworks.

\section{Preliminary}\label{sec:pre}
We first establish the formal definitions and notations that form the foundation of our approach. 
\begin{definition}[Multitask Learning] 
\label{def:mtl}
Consider a set of tasks $\mathcal{T} = \{T_i \mid i \in [n]\}$, where $[n] = \{1, 2, \ldots, n\}$ and each task $T_i$ is associated with a learning objective $L_i(\phi, \theta_i)$. Here, $\phi \in \mathbb{R}^p$ represents the shared parameters across all tasks, and $\theta_i \in \mathbb{R}^{p_i}$ denotes the task-specific parameters for task $T_i$. The multitask learning objective is to jointly minimize a weighted combination of individual task losses:
{\small
\begin{equation}
\phi^*, \{\theta_i^*\}_{i=1}^n = \arg \min_{\phi, \{\theta_i\}_{i=1}^n} \sum_{i=1}^{n} \lambda_i L_i(\phi, \theta_i),
\end{equation}
}

where $\{\lambda_i > 0\}_{i=1}^n$ are task-specific weights that balance the contribution of each task.
\end{definition}

While conventional multitask learning directly optimizes the shared and task-specific parameters through the combined objective, our approach exploits the inherent relationships between tasks during the training process. Rather than traing all tasks jointly within a single optimization framework, we organize them into groups based on their mutual influences, leading to the concept of task grouping.

\begin{definition}{(Task Grouping)}
\label{tg}
Let $\mathcal{T} = \{T_i \mid i \in [n]\}$ denote the set of tasks, and $G = \{G_j \mid j \in [m]\}$ represent the set of task groups. Task grouping aims to establish a mapping based on task relationships such that for every task $T_i$, there exists a group $G_j$ to which $T_i$ is assigned, ensuring the inclusion of at least one task in each group and resulting in the best performance for each task.
\end{definition}

After establishing the task groups, we optimize parameters separately within each group:
\begin{small}
\begin{equation}\label{eq: obj tg}
\min_{\phi_j, \{\theta_i\}_{i \in G_j}} \sum_{i \in G_j} \lambda_i L_i(\phi_j, \theta_i),\ \forall G_j \in G
\end{equation}
\end{small}

\vspace{-2mm}
where $\phi_j$ represents the shared parameters specific to group $G_j$. This group-wise optimization approach enhances both computational efficiency and learning effectiveness by allowing each group to focus on a coherent set of related tasks.

\section{Method}\label{sec:method}
To infer task groupings for subsequent optimization processes in Equation \eqref{eq: obj tg}, we introduce a methodology for constructing transfer gains, as elucidated in Section \ref{sec: tac}, demonstrating its efficacy in yielding theoretical outcomes without relying on underlying assumptions. Subsequently, we propose a versatile mathematical programming framework in Section \ref{sec: tga} that flexibly accommodates various budget constraints. This formulation is instrumental in deriving the outcomes of task grouping. Furthermore, in Section \ref{sec: aa}, we conduct a detailed analysis of the computational complexity associated with collecting transfer gains, in comparison to TAG \cite{fifty2021efficiently}.

\subsection{Assumption-Free Transfer Gain}\label{sec: tac}

In this subsection, we introduce the pivotal concept of proposed transfer gain in our method.
\begin{definition}{(Transfer Gain)}\label{def: ta}
For tasks $T_i \neq T_j$, the \emph{task transfer gain} from $T_i$ to $T_j$ at training step $t$ is characterized by: 
\begin{small}
    \begin{equation}\label{eq:task aff per step}
\mathcal{S}_{i \rightarrow j}^{t} = 1 - \frac{L_{j}\left(\phi_{\{i,j\}}^{t+1}, \theta_{j}^{t+1}\right)}{L_{j}\left(\phi_{\{j\}}^{t+1}, \theta_{j}^{t+1}\right)}.
\end{equation} 
\end{small}

\vspace{-2mm}
In this equation, $L_j$ represents a task-specific metric, such as the loss function or validation accuracy. $\phi_{\{i,j\}}^{t+1}$ and $\theta_{j}^{t+1}$ represent the model parameters trained by $T_i$ and $T_j$ at the subsequent training iteration.
We then define the \emph{group transfer gain} from any $A \subseteq \mathcal{T}$ to task $T_j$ as: {\small
\begin{equation}\label{eq:group task aff per step}
\mathcal{S}_{A \rightarrow j}^{t} = 1 - \frac{L_{j}\left(\phi_{A\cup\{j\}}^{t+1}, \theta_{j}^{t+1}\right)}{L_{j}\left(\phi_{\{j\}}^{t+1}, \theta_{j}^{t+1}\right)}.
\end{equation} }

\vspace{-2mm}
Furthermore, we extend this concept to define the \emph{group transfer gain} from any $A \subseteq \mathcal{T}$ to $B \subseteq \mathcal{T}$ as: 
{\small
\begin{equation}\label{eq:group2group task aff per step}
\mathcal{S}_{A \rightarrow B}^{t} = \sum_{j \in B} \mathcal{S}_{A \rightarrow j}^{t},
\end{equation}}

\vspace{-2mm}
which allows us to measure the collective transfer of knowledge between groups of tasks. 
\end{definition}

While our formulation, as presented in Equation \eqref{eq:task aff per step}, bears a formal resemblance to TAG \cite{fifty2021efficiently}, there are essential differences between the two. We will elucidate these distinctions and demonstrate the superior advantages of our approach in the ensuing discussion.

\textbf{First}, \cite{fifty2021efficiently} defines task affinity as $\mathcal{Z}_{i \rightarrow j}^{t} = 1 - \frac{L_{j}\left(\phi_{\{i\}}^{t+1}, \theta_{j}^{t}\right)}{L_{j}\left(\phi^{t}, \theta_{j}^{t}\right)}$ with respect to the loss function, reflecting the effects of training $T_i$ on $T_j$, while $\mathcal{S}^t_{i \rightarrow j}$ measures the effects of training $T_i$ on \textit{training} $T_j$. This substantial distinction allows us to establish the relationship between loss decrease and the value of $\mathcal{S}^t_{i \rightarrow j}$ without additional assumptions. Specifically, if $\mathcal{S}^t_{i \rightarrow j} > \mathcal{S}^t_{k \rightarrow j}$, then training $\{T_i, T_j\}$ results in a greater loss decrease than training $\{T_k, T_j\}$, as summarized in the following observation:
\begin{observation}\label{obs: loss dec}
If \(\mathcal{S}_{A_1 \rightarrow i}^{t} > \mathcal{S}_{A_2 \rightarrow i}^{t}\), then training task group \(A_1\) induces a larger loss decrease than \(A_2\) for task \(i\).
\end{observation}
In contrast, \cite{fifty2021efficiently} introduces restrictive constraints, such as strong convexity on loss functions, to enforce this relationship.
\textbf{Second}, in \cite{fifty2021efficiently}, group transfer gain is formulated as $\mathcal{Z}_{\{j,k\} \rightarrow i}^{t} = \frac{1}{2}\left(\mathcal{Z}_{j \rightarrow i}^{t} + \mathcal{Z}_{k \rightarrow i}^{t}\right)$. This formulation directly defines the group transfer gain at the task level, lacking theoretical guarantees of effectiveness. In this work, we establish the connection between task and group transfer gain in Proposition \ref{propos: linear prop}, providing both theoretical advantages and valid empirical operations at the implementation level.

\begin{proposition} \label{propos: linear prop}
Consider a multi-task learning setup with shared parameters $\phi \in \mathbb{R}^d$ and task-specific parameters $\theta_k$ for each task $T_k \in \mathcal{T}$. Let $L_k(\phi, \theta_k)$ be the loss function for task $T_k$. Suppose the model parameters are updated from $(\phi^t, \{\theta_k^t\}_{k \in \mathcal{T}})$ to $(\phi^{t+1}, \{\theta_k^{t+1}\}_{k \in \mathcal{T}})$ using a single step of gradient descent with learning rate $\eta_t > 0$.
Assume the following conditions hold:
\vspace{-2mm}
\begin{enumerate}
    \item For all tasks $k \in \mathcal{T}$, the loss function $L_k(\phi, \theta_k)$ is $l$-Lipschitz with respect to $\phi$ for any fixed $\theta_k$, and differentiable with respect to $\phi$. This implies that $||\nabla_\phi L_k(\phi, \theta_k)|| \le l$ for all $\phi, \theta_k$.
    \vspace{-3mm}
    \item There exists a constant $C > 0$ such that $L_j\left(\phi_{\{j\}}^{t+1}, \theta_{j}^{t+1}\right) \ge C$.
\end{enumerate}
\vspace{-3mm}

Let $\mathcal{S}_{i \rightarrow j}^{t}$ and  $\mathcal{S}_{A \rightarrow j}^{t}$ be the task transfer gain and group transfer gain as defined in Equation \ref{eq:task aff per step} and \ref{eq:group task aff per step}.
Then, we have the following bound on the absolute difference between the group transfer gain and the average of individual task transfer gains:
{\small
\begin{equation} \label{eq:prop_bound_formal}
\left|\mathcal{S}_{A \rightarrow j}^{t} - \frac{1}{|A|}\sum_{i \in A} \mathcal{S}_{i \rightarrow j}^{t}\right| \le \frac{\eta_t(1+|A|)l^2}{C}.
\end{equation}}
\end{proposition}

\begin{remark}[Practical Implications of the Bound's Magnitude] \label{rem: bound_practicality}
In practical training scenarios, several factors contribute to making this upper bound small, supporting the notion that the group transfer gain is often well-approximated by the average of individual transfer gains:
    \textbf{Learning Rate:} Standard optimization practices for neural networks employ relatively small learning rates, and often use schedules that decrease $\eta_t$ over time. A small $\eta_t$ directly scales down the entire bound;
    \textbf{Group Size:} The number of source tasks included in a group for simultaneous training is typically constrained by available computational resources, such as GPU memory and processing power;
    \textbf{Loss Lower Bound:} The condition $L_j(\phi_{\{j\}}^{t+1}, \theta_j^{t+1}) \ge C > 0$ is often met or promoted through standard training techniques. Regularization methods (e.g., L2 regularization, dropout) and early stopping prevent the loss function from converging to zero;
    \textbf{Lipschitz Constant:} While the Lipschitz constant $l$ depends on the choice of model architecture and loss function, for many commonly used models and loss functions (e.g., truncated gradients for bounded inputs), $l$ is finite.

Collectively, the use of small learning rates, practical limits on group size, and techniques that ensure a positive lower bound on the loss contribute to making the upper bound $\frac{\eta_t(1+|A|)l^2}{C}$ small. 
\end{remark}


Based on these findings, $\mathcal{S}_{i \rightarrow j}^{t}$ serves as a measure of intrinsic signals that elucidate the relationships between tasks during training. Furthermore, we introduce the concepts of cumulative transfer gain and group transfer throughout the training process. These are defined as follows: 
\vspace{-1mm}
{\small
\begin{equation}\label{eq: cumulative transfer}
\mathcal{S}_{i \rightarrow j}^{} = \frac{1}{T}{\sum_{t=1}^{T}\mathcal{S}_{i \rightarrow j}^{t}},\qquad\ \mathcal{S}_{A \rightarrow j}=\frac{1}{|A|}\sum_{i\in A}\mathcal{S}_{i \rightarrow j}.
\end{equation}}

\vspace{-2mm}
These metrics quantify the impact of training group $A$ on task $j$ over the entire training period.

\subsection{Generic and Flexible Task Grouping Framework}\label{sec: tga}
After obtaining the cumulative transfer gain $\mathcal{S}$ from Equation~\eqref{eq: cumulative transfer}, the remaining challenge is to accurately determine the task grouping results. Practical applications often require customization based on various constraints, such as resource limitations. Therefore, it is crucial to develop a general and flexible task grouping framework. Unlike previous approaches \cite{zamir2018taskonomy, fifty2021efficiently, standley2020tasks}, which utilize Binary Integer Programming and Branch \& Bound methods, we propose a mathematical programming framework that can be transformed into Mixed Integer Programming.
This involves setting binary variables $X_{ij}$, where $i \in [n]$ and $j \in [m]$, with $X_{ij} = 1$ indicating the assignment of task $i$ to group $j$, and $X_{ij} = 0$ otherwise. Notably, $X_{\cdot j} \in \mathbb{R}^n$ represents the $j$-th column of $X$. The vector $\mathbf{1}$ is composed entirely of ones with an adaptable dimension. The element $B_{ij}$ in $B \in \mathbb{R}^{n \times m}$ denotes the budget of task $T_i$ assigned to group $G_j$. The vector $\mathbf{b} \in \mathbb{R}^m$ represents the maximum budget for each group. The symbol “$\odot$” signifies the element-wise product between matrices, resulting in the following formulation:
\vspace{-3mm}
\begin{small}
\begin{equation}\label{eq: formulation}
\begin{aligned}
\max_{X} \quad & \sum_{j=1}^{m}\frac{1}{\mathbf{1}^\top X_{\cdot j}}X_{\cdot j}^\top \mathcal{S}X_{\cdot j} \\
\text{s.t.} \quad 
   &{X^\top \mathbf{1}\ge \mathbf{1}},\quad X \mathbf{1}\ge \mathbf{1}, \quad (B\odot X)^\top\mathbf{1}\le \mathbf{b}\\ 
    &||X_{\cdot j_1}-X_{\cdot j_2}||^2\ge 1\text{ for } j_1\neq j_2, \quad {X\in\{0,1\}^{n\times m}}.
\end{aligned}
\end{equation}    
\end{small}
\vspace{-3mm}

The primary objective of the task grouping problem is to compute the aggregate impact of training each group individually. The quadratic form $X_{\cdot j}^\top \mathcal{S} X_{\cdot j}$ arises from the approximation stated in Proposition~\ref{propos: linear prop}. Several key constraints are imposed to ensure a meaningful solution. The constraint $X^\top \mathbf{1} \geq \mathbf{1}$ is established to guarantee that each group contains at least one task, while the constraint $X \mathbf{1} \geq \mathbf{1}$ ensures that all tasks are incorporated into the grouping outcomes. Additionally, diverse budgetary constraints can be introduced by incorporating the inequality $(B \odot X)^\top \mathbf{1} \leq \mathbf{b}$.

Moreover, it is imperative that the resulting groups remain distinct, and this distinctiveness is ensured by the condition $\|X_{\cdot j_1} - X_{\cdot j_2}\|^2 \geq 1$. Remarkably, recent advancements in powerful mathematical programming solvers have significantly enhanced the efficiency of resolving these problems through appropriate transformations, as elucidated in Appendix \ref{app: transformation}.

\subsection{Analysis of Computation Complexity}\label{sec: aa}
In this section, we discuss the computational complexity of the proposed method for collecting transfer gains and compare it to the most closely related work, TAG \cite{fifty2021efficiently}. We begin by defining the relevant notations: For each task $T_i$ with neural network capacity $\mathcal{C}_i$, the feed-forward and backward computational costs of the neural networks are denoted as $\mathcal{F}_i$ and $\mathcal{B}_i$, respectively. The average computation costs and capacities for these tasks are represented by $\mathcal{F}$, $\mathcal{B}$, and $\mathcal{C}$, where $\mathcal{F} = \frac{1}{n} \sum_{i=1}^n \mathcal{F}_i$, $\mathcal{B} = \frac{1}{n} \sum_{i=1}^n \mathcal{B}_i$, and $\mathcal{C} = \frac{1}{n} \sum_{i=1}^n \mathcal{C}_i$. Based on these notations, the computation costs for TAG and our proposed method are $(n^2+n)\mathcal{F} + n\mathcal{B} + n\mathcal{C}$ and $(n^2+2n)\mathcal{F} + n\mathcal{B} + (n^2+n)\mathcal{C}$, respectively. Both methods have the same order of computational complexity with respect to the number of tasks.
While our method involves an additional $n$ feed-forward computations and $n^2$ weight assignment operations for the neural networks, it is important to emphasize that feed-forward and weight assignment costs are typically much smaller than backward pass costs in practice. Consequently, the impact of these additional feed-forward computations is minimal, especially when considering the theoretical advantages our method offers.

To further enhance computational efficiency when the number of tasks is large, we propose two practical approaches for collecting transfer gains:
(1) \textbf{Sampling strategy}: We implement a sampling-based approach for collecting transfer gains. We define a random variable $T$ that follows a uniform distribution, denoted as $T \sim \text{Unif}\{1, 2, \ldots, n\}$. A subset of tasks of size $T$ is then randomly selected, and transfer gains are gathered solely from this subset. This sampling strategy significantly reduces the computational cost of our method to $\frac{n^2+6n+2}{3}\mathcal{F}+n\mathcal{B}+\frac{(n+1)(n+5)}{6}\mathcal{C}$ in expectation, which is substantially lower than that of TAG. 
(2) \textbf{Lazy collection strategy}: Employing a lazy collection strategy for transfer gains maintains performance while reducing computational costs. The analysis reveals that cumulative transfer gain varies across different training phases, yielding diverse outcomes. These insights allow for substantial reductions in computational costs.
Detailed analysis and results are provided in Section \ref{sec: efficiency impr.}.
\begin{figure}[htbp!]
     \centering
     \includegraphics[width=\textwidth]{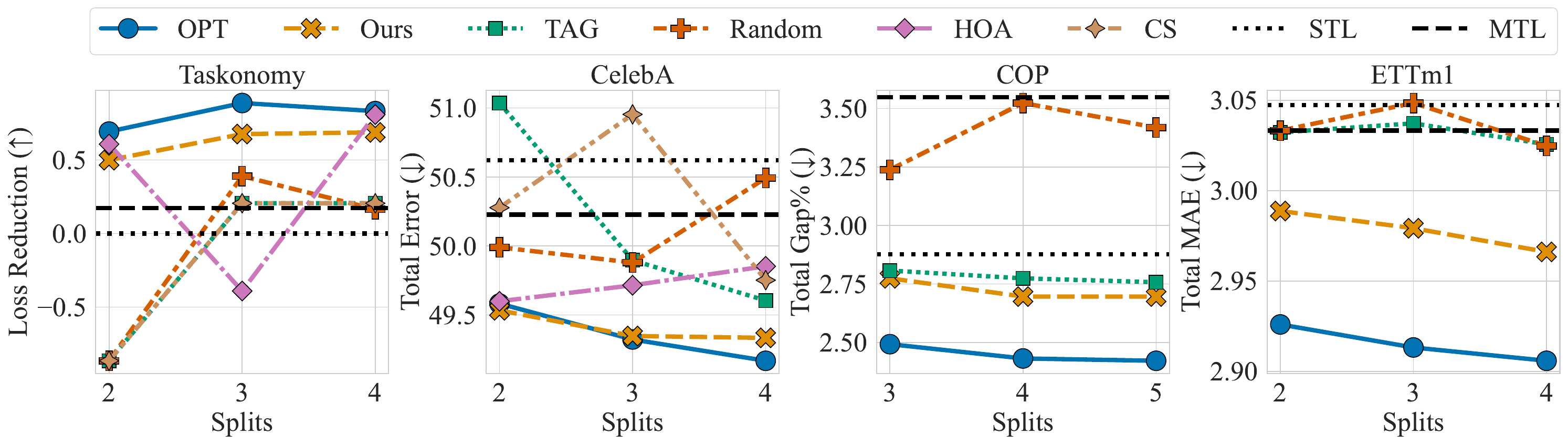}
    \caption{\label{fig: all result}
    Performance demonstration across grouping methods on each dataset. This figure presents the results in loss reduction ($\uparrow$) for Taskonomy, total test error ($\downarrow$) for CelebA, total optimality gap ($\downarrow$) and total MAE ($\downarrow$) for ETTm1, segmented by various data splits. The dotted and dashed horizontal lines indicate the Single Task Learning (STL) and the best Multi-Task Learning (MTL) benchmarks. 
    }
    \vspace{-5mm}
\end{figure}


\section{Experiments}\label{submission}
In this section, we present the experimental evaluation of our task grouping approach, demonstrating its effectiveness across diverse domains such as computer vision, combinatorial optimization, and time series analysis. 
Due to space constraints, only the primary experimental settings and results are included. Comprehensive details about experimental setups, analysis of group results, and digital tables can be found in Appendix \ref{app: detailed result}.

\subsection{Experimental Setups}\label{sec: exp setup}
\textbf{Datasets and Metrics.} 
Our experiments are designed to validate the versatility and superiority of our method across four datasets:
    (1) Taskonomy \cite{zamir2018taskonomy} and (2) CelebA: These are classical computer vision datasets used in previous task grouping methods. Following the settings in TAG \cite{fifty2021efficiently}, we selected five tasks from Taskonomy and nine tasks from CelebA for our experiments. The performance of each method is evaluated by the total loss reduction compared with single-task learning (STL) for Taskonomy and the classification error rates for CelebA across all tasks. To ensure experimental consistency, we adhere to the network architecture and training hyperparameters specified in TAG \cite{fifty2021efficiently};
    (3) Combinatorial Optimization Benchmarks \cite{wang2023efficient}: We test six tasks: TSP20, TSP50, CVRP20, CVRP50, OP20, and OP50, representing various scales of the Traveling Salesman Problem (TSP), Capacitated Vehicle Routing Problem (CVRP), and Orienteering Problem (OP). Performance is evaluated based on the average optimality gap, mathematically defined as: $
    \text{Gap}\% = \frac{1}{N} \sum_{i=1}^{N}\left(1 - \frac{\text{solver}(\mathcal{I}_i)}{\text{gt}(\mathcal{I}_i)}\right) \times 100,
    $ evaluated over $N = 10,000$ instances for each task to measure the solution's deviation from the ground truth obtained from Gurobi \cite{gurobi}. The neural solver used for these tasks is the POMO framework \cite{kwon2020pomo}, noted for its effectiveness in addressing combinatorial optimization problems;
    (4) ETTm1 \cite{wu2021autoformer}: This is an electric load dataset with seven time series. Effectiveness is assessed using the mean absolute error's (MAE) relative reduction as the evaluation metric. We employ the model architecture based on the AutoFormer framework \cite{wu2021autoformer}, specifically designed for the intricacies and predictive challenges of multivariate time series data.

\textbf{Baselines } In general, our experimental evaluation involves a comprehensive comparison against a range of established methods: {(1)} Single Task Learning (STL); {(2)} MTL methods: We consider a variety of MTL methods that employ different strategies for joint task learning, including: Naive-MTL, Bandit-MTL \citep{mao2021banditmtl}, PCGrad \citep{yu2020gradient}, Nash-MTL \citep{navon2022multi}, Uncertainty-Weighting (UW) \citep{kendall2018multi} and LinearScale; {(3)} Task grouping methods: Random policy by which tasks are grouped randomly and results are taken the average for 10 repeats; Optimal policy which is obtained by enumeration; TAG \cite{fifty2021efficiently}, a known task grouping SOTA method to group tasks based on their affinity.
Additionally, we include extra high order approximation (HOA) \cite{standley2020tasks} and cosine similarity (CS) in Taskonomy and CelebA and MTG \cite{song2022efficient} in ETTm1.

\subsection{Main Results}\label{sec: main res}
In this subsection, we present the empirical evaluation of our task grouping approach across four diverse domains: computer vision (Taskonomy and CelebA datasets), combinatorial optimization problems (COP), and time series forecasting (ETTm1 dataset). For brevity, we present the most significant findings from Figure \ref{fig: all result} here, while comprehensive digital tables with detailed results are provided in the Appendix: Taskonomy results in Section \ref{app: taskonomy res}, CelebA results in Section \ref{app: celeba res}, combinatorial optimization results in Section \ref{app: cop res}, and time series forecasting results in Section \ref{app: ts res}.

\textbf{Results on Taskonomy.} 
In Figure \ref{fig: all result}, the leftmost figure for the Taskonomy dataset focuses on evaluating the loss reduction compared to STL, where higher values indicate better performance. Results show that our method consistently outperforms the STL and MTL benchmarks across all splits, demonstrating robust and reliable performance compared to traditional learning approaches. Furthermore, our method shows a clear trend of improving performance with an increasing number of splits, indicating that finer granularity in task grouping leads to better results. While HOA performs well in splits 2 and 4, it shows a significant drop in performance at split 3, highlighting its lack of stability. Although our method occasionally performs slightly worse than HOA in certain splits, it provides superior overall stability and consistency. Given that HOA's high computational cost makes it less practical, the proposed method stands out as the best choice for achieving a balance between performance, stability, and computational efficiency.

\textbf{Results on CelebA.}
The second-left figure in Figure \ref{fig: all result} provides a comprehensive evaluation of various baselines in terms of the total error metric for the CelebA dataset, where lower values indicate better performance. It is observed that MTL exhibits superior performance compared to STL, and most grouping methods result in a further reduction in total error. This suggests the presence of complex underlying relationships between the tasks in this dataset, indicating that our grouping methods can identify specific combinations that enhance overall performance. Our method consistently outperforms all baselines in terms of total error across all splits, with the only exception being the optimal grouping strategy found in TAG \cite{fifty2021efficiently}. Additionally, our method effectively leverages the granularity afforded by increased splits, demonstrating a decrease in total error.

\textbf{Results on COP.}
In the comparative analysis presented in Figure \ref{fig: all result}, STL demonstrates a robust baseline, outperforming MTL methods with respect to the total gap metric. Within the domain of task grouping methods, both TAG and our proposed method have shown the capability to surpass the STL baseline in certain aspects. Notably, our method consistently achieves the best performance among non-optimal baselines across each grouping strategy, indicating its efficacy in handling multiple related tasks simultaneously. As we consider the performance trends across different task groupings, it is observed that the efficacy of Optimal, TAG, and our method improves as the splits become larger. This trend suggests that the tasks within the COP benchmark exhibit high positive transfer potential. Detailed grouping analysis can be seen in Appendix \ref{app: cop res}, where our method achieves logical groupings of tasks, pairing the same types of COPs together: (TSP20, TSP50), (CVRP20, CVRP50), and (OP20, OP50) when splitting these tasks into three groups. This reflects an intuitive understanding that tasks of the same types benefit from being trained together.

\textbf{Results on ETTm1.}
The rightmost figure in Figure \ref{fig: all result} evaluates the total MAE for the ETTm1 dataset, where lower values indicate better performance. It shows that our method consistently outperforms all non-optimal benchmarks across all splits. Moreover, it demonstrates a trend of improving performance with an increasing number of splits, highlighting its robustness and reliability in minimizing MAE. However, TAG shows performance comparable to the Random baseline in most splits and exceeds both STL and MTL benchmarks only at the 4-split level. This variability in TAG's performance suggests that it struggles to identify and leverage the intricate connections between time series tasks. In contrast, the superior performance of our method highlights its capability to effectively uncover and utilize the intrinsic relationships between time series tasks, leading to the best overall results.
    
\subsection{Task Grouping with Constraints}\label{sec: introduce constraints}
In practical scenarios, constraints such as limited computational resources, data availability, and group size requirements are common. Tasks often involve varying data acquisition costs, necessitating budget management. In distributed learning, where resources are distributed across nodes, complying with group size limits is crucial. Our mathematical programming approach, as outlined in Formulation \ref{eq: formulation}, effectively addresses the task grouping challenge under these constraints by incorporating customized limitations to meet the practical demands of real-world applications.
In this section, we address the constraint that group sizes must fall within a specified range, represented as: $
\mathbf{m_1} \le X^T \mathbf{1} \le \mathbf{m_2},$ where $\mathbf{m_1}, \mathbf{m_2} \in \mathbb{R}^m$, using element-wise comparison. Our experiments in computer vision, COP, and time series applications demonstrate the model’s ability to adhere to these size constraints. We benchmark our method against random sampling, conducted ten times under the same constraints, with the experimental setup detailed subsequently.
\begin{figure}[t!]
     \centering
    \begin{subfigure}{.49\textwidth}
         \centering
    \includegraphics[width=\textwidth]{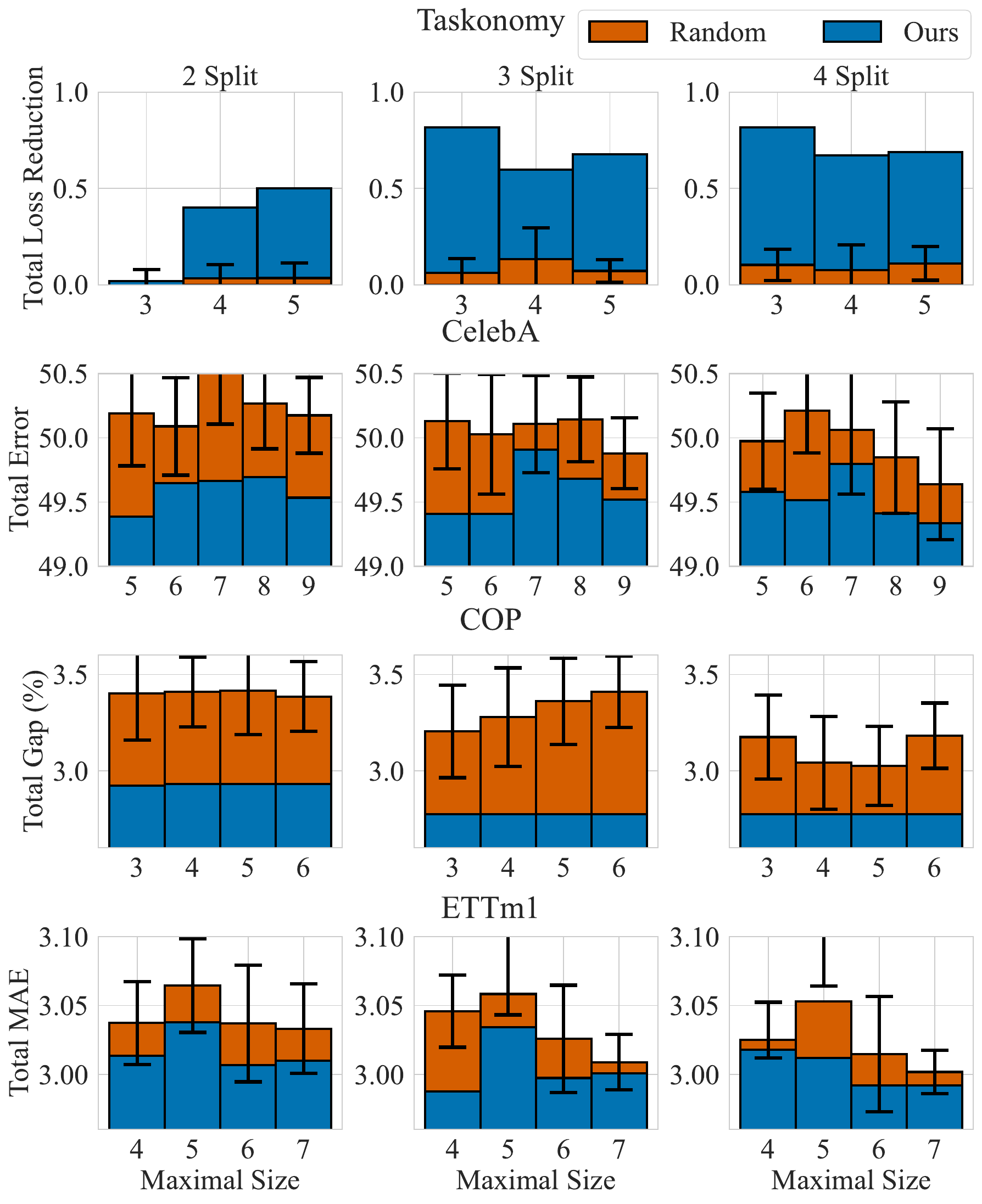}
         \caption{Maximal Group Constraint\label{max constr}}
     \end{subfigure}
        \begin{subfigure}{.49\textwidth}
         \centering
    \includegraphics[width=\textwidth]{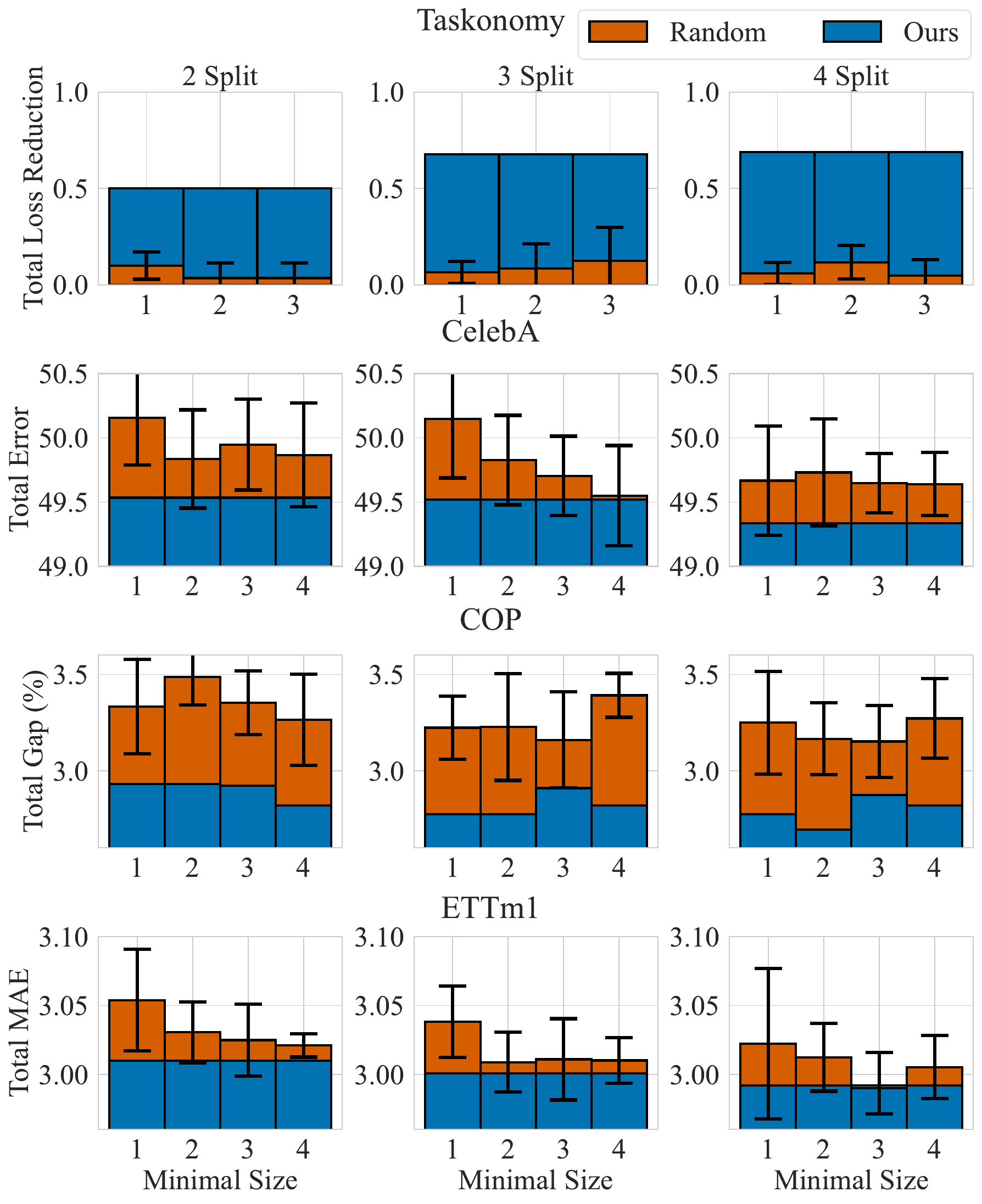}
         \caption{Minimal Group Constraint\label{min constr}}
     \end{subfigure}
\caption{\label{fig:result constr}
Comparative Performance under Maximum and Minimum Group Size Constraints. The figure delineates the performance of our task grouping method against random policy. Metrics such as loss reduction $(\uparrow)$ for Taskonomy dataset, total error $(\downarrow)$ for the CelebA dataset, Total Gap $(\downarrow)$ for combinatorial optimization problems (COP), and Total Mean Absolute Error (MAE) $(\downarrow)$ for time series forecasting tasks are evaluated across a range of group sizes, illustrating the adaptability of our method to both maximal and minimal size constraints.}
\vspace{-7mm}
\end{figure}

    \textbf{Maximum Group Size Constraint.} 
    For this constraint, we define $\mathbf{m_1}$ as $\mathbf{1}$ and $\mathbf{m_2}$ as $M\mathbf{1}$, where $M$ represents the maximum permissible group size, dictated by memory limitations. For the Taskonomy dataset containing five tasks, we select $M$ from $\{3, 4, 5\}$. In the context of the CelebA dataset, which comprises nine tasks, we set $M$ within the range $\{5, 6, 7, 8, 9\}$. For the COP benchmark involving six tasks, $M$ is selected from $\{3, 4, 5, 6\}$. For time series tasks comprising seven tasks, we choose $M$ from $\{4, 5, 6, 7\}$. The experimental results, as illustrated in Figure \ref{max constr}, exhibit a uniform trend across varying cases and group sizes. For example, in the Taskonomy and CelebA datasets, our method consistently exhibits superior performance compared to the random policy. This is evident in the progressively larger loss reduction and lower total error rates as the maximum group size increases. Similarly, in the COP benchmarks and time series tasks with their respective maximum group size constraints, our method maintains or even enhances its performance. This consistent trend, observed across various maximum group sizes and tasks, underscores the robustness and adaptability of our approach in accommodating changing task constraints.

    \textbf{Minimum Group Size Constraint.} 
    In this case, the vectors $\mathbf{m_1}$ and $\mathbf{m_2}$ are defined as $m\mathbf{1}$ and $M_{\text{max}}\mathbf{1}$, respectively, where $M_{\text{max}}$ denotes the total number of tasks specific to each case: five for the Taskonomy dataset, nine for the CelebA dataset, six for COP, and seven for time series tasks. This formulation ensures that device utilization at each node surpasses a certain threshold, thereby guaranteeing efficient usage. For the Taskonomy dataset, we select $m$ from $\{1, 2, 3\}$. For the remaining datasets, $m$ is chosen from the set $\{1, 2, 3, 4\}$.
    Results are presented in Figure \ref{min constr}. In the CV datasets, as the minimum group size increases, our method consistently demonstrates superior performance compared to the random policy, indicating improved efficiency in handling larger group sizes. Similarly, in the COP benchmarks, the Total Gap percentage decreases as the group size grows, highlighting our method's effectiveness under tighter constraints. Although the performance of our method in the time series tasks under minimum group size constraints does not match the levels achieved in the CV and COP benchmarks, it still significantly outperforms the random policy.
    
   \subsection{Improve Training Efficiency for Task Grouping}\label{sec: efficiency impr.}
\begin{figure*}[t!]
    \centering
    \begin{minipage}[t]{0.5\textwidth}
        \centering
        \captionsetup{type=figure}
        \includegraphics[width=.96\linewidth]{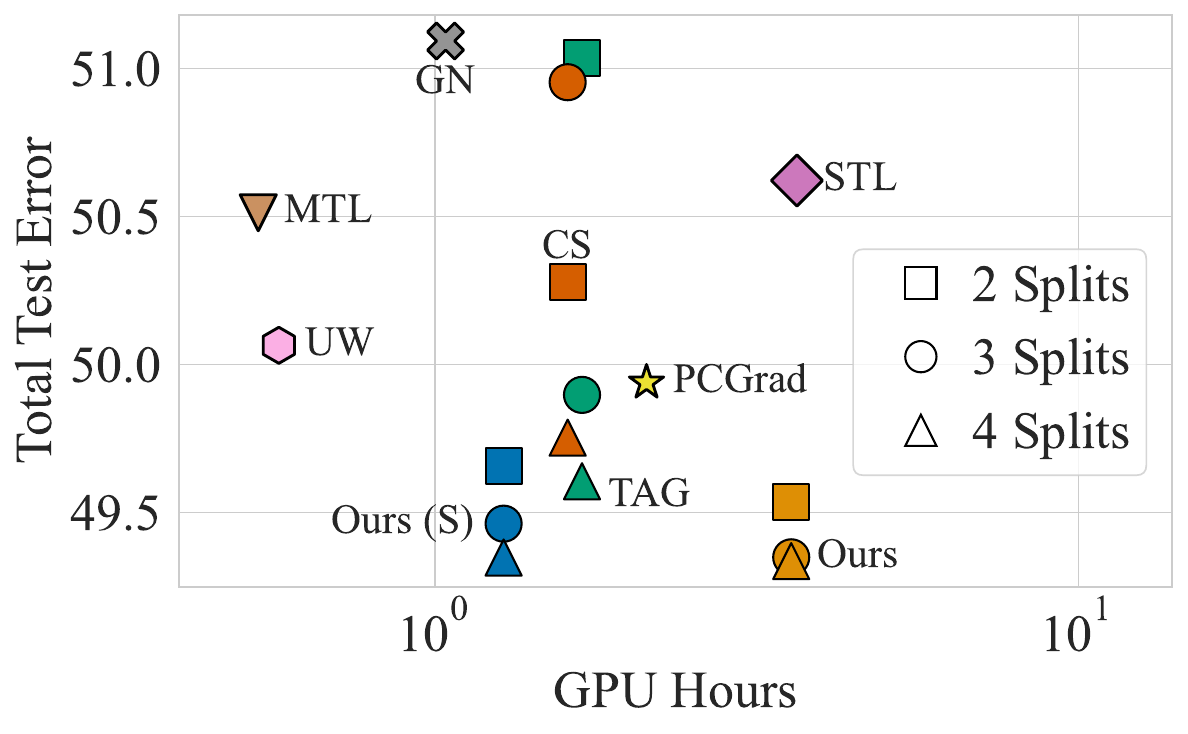}
         \caption{Average classification error for 2, 3, and 4-split task groupings for the subset of 9 tasks in CelebA, compared across various methods (Ours (S), Ours, TAG, CS, STL, MTL, UW, GN, PCGrad) versus GPU hours.} 
        \label{fig:celeba_sampling} 
    \end{minipage}
        \hfill 
    \begin{minipage}[t]{0.48\textwidth} 
        \centering
        \captionsetup{type=table}
        \footnotesize
        \setlength{\tabcolsep}{0.3em} 
        \renewcommand{\arraystretch}{1.2} 
        \scalebox{.89}{
        \begin{tabular}{c|cc|cc}
        \toprule
        \multirow{2}{*}{Freq.} & \multicolumn{2}{c|}{Relative Improvement $(\uparrow)$} & \multicolumn{2}{c}{Relative Speedup $(\uparrow)$} \\
        \cmidrule(lr){2-3} \cmidrule(lr){4-5}
        & CelebA & ETTm1 & CelebA & ETTm1 \\
        \midrule
        1 & 3.50\% & 5.76\% & 1.00 & 1.00 \\
        5 & 3.53\% & 9.48\% & 4.25 & 4.62 \\
        10 & 3.94\% & 9.20\% & 7.18 & 8.45 \\
        25 & 4.60\% & 8.75\% & 12.16 & 16.64 \\
        50 & 5.13\% & 8.75\% & 15.95 & 24.94 \\
        100 & 1.83\% & 6.70\% & 15.95 & 33.24 \\
        200 & 3.06\% & 5.87\% & 18.62 & 39.16 \\
        \bottomrule
        \end{tabular}}
        \vspace{2.4mm}
        \caption{ \label{tab:combined_interval} Test errors across different frequencies of transfer gain collection, compared to test errors from random groupings, with speedup evaluated based on the computation of transfer gains at each step. 
        } 
    \end{minipage}
    \vspace{-5mm}
\end{figure*}

As discussed in Section \ref{sec: aa}, computational complexity becomes a significant concern when applying our task grouping method to scenarios with a large number of tasks. Here, we empirically evaluate the two efficiency-enhancing strategies proposed earlier: the sampling strategy and the lazy collection strategy. Our experiments demonstrate that these approaches substantially reduce computational overhead while maintaining the quality of task groupings.
\vspace{-3mm}
\paragraph{Sampling Strategy.}
We evaluate our sampling strategy on the CelebA dataset (9 tasks) by comparing four grouping methods in Figure \ref{fig:celeba_sampling}: our original approach, TAG, CS, and our sampling-enhanced method (Ours (S)). We also benchmark against five non-grouping methods: STL, MTL, UW, GradNorm, and PCGrad. The analysis focuses on both computational efficiency and performance effectiveness.
Our original method achieves the lowest total test error across all splits, demonstrating superior performance but at the cost of increased runtime. By incorporating the sampling strategy, Ours (S) significantly reduces computational overhead while maintaining comparable performance. Notably, Ours (S) outperforms TAG in both accuracy and efficiency, validating its effectiveness.
\vspace{-3mm}
\paragraph{Lazy Collection Strategy.}
We analyze how the frequency of transfer gain collection affects performance and computational efficiency across two datasets: CelebA and ETTm1. 
Table \ref{tab:combined_interval} presents relative improvements and computational speedups achieved with collection intervals ranging from 1 to 200 steps. For the CelebA dataset, relative improvement actually increases as collection frequency decreases (up to an interval of 50 steps). This counter-intuitive result suggests that less frequent collection may reduce estimation noise, leading to more robust task groupings. Beyond the 50-step threshold, performance begins to decline as transfer gain estimates become increasingly inaccurate due to insufficient sampling. 
The ETTm1 dataset exhibits a similar pattern, with optimal performance occurring at moderate collection intervals. These results demonstrate that our method can achieve substantial speedups (10\textendash{}15×) by employing a lazy collection strategy with intervals of 10\textendash{}50 steps, while simultaneously maintaining or even enhancing performance. This finding has significant practical implications for deploying our method in compute-constrained environments.

\section{Conclusions and Limitations}
In conclusion, our work presents a novel approach to task grouping in Multi-Task Learning (MTL), marking a significant advancement over existing methods. Our approach features two innovations: a robust transfer gains construction that operates without restrictive assumptions, and a flexible mathematical programming formulation tailored for task grouping challenges. Empirical results demonstrate our method's superiority, showing improved performance, flexibility, and efficiency in real-world settings, thus enhancing MTL models' applicability in diverse and resource-limited environments.
Despite its promising results, the scalability of our method with an extremely large number of tasks is yet to be determined. Moreover, while our MIP framework is versatile, it currently focuses on knapsack constraints. Introducing additional constraint types could broaden its applicability and better address complex real-world scenarios.




\bibliographystyle{unsrt}
\bibliography{ref}

\begin{thebibliography}{10}

\bibitem{caruana1997multitask}
Rich Caruana.
\newblock Multitask learning.
\newblock {\em Machine learning}, 28:41--75, 1997.

\bibitem{zhang2021survey}
Yu~Zhang and Qiang Yang.
\newblock A survey on multi-task learning.
\newblock {\em IEEE Transactions on Knowledge and Data Engineering}, 34(12):5586--5609, 2021.

\bibitem{vandenhende2021multi}
Simon Vandenhende, Stamatios Georgoulis, Wouter Van~Gansbeke, Marc Proesmans, Dengxin Dai, and Luc Van~Gool.
\newblock Multi-task learning for dense prediction tasks: A survey.
\newblock {\em IEEE transactions on pattern analysis and machine intelligence}, 44(7):3614--3633, 2021.

\bibitem{standley2020tasks}
Trevor Standley, Amir Zamir, Dawn Chen, Leonidas Guibas, Jitendra Malik, and Silvio Savarese.
\newblock Which tasks should be learned together in multi-task learning?
\newblock In {\em ICML}, pages 9120--9132, 2020.

\bibitem{fifty2021efficiently}
Chris Fifty, Ehsan Amid, Zhe Zhao, Tianhe Yu, Rohan Anil, and Chelsea Finn.
\newblock Efficiently identifying task groupings for multi-task learning.
\newblock {\em Advances in Neural Information Processing Systems}, 34:27503--27516, 2021.

\bibitem{song2022efficient}
Xiaozhuang Song, Shun Zheng, Wei Cao, James Yu, and Jiang Bian.
\newblock Efficient and effective multi-task grouping via meta learning on task combinations.
\newblock {\em Advances in Neural Information Processing Systems}, 35:37647--37659, 2022.

\bibitem{sherif2023stg}
Ammar Sherif, Abubakar Abid, Mustafa Elattar, and Mohamed ElHelw.
\newblock Stg-mtl: Scalable task grouping for multi-task learning using data map.
\newblock {\em arXiv preprint arXiv:2307.03374}, 2023.

\bibitem{zhang2022survey}
Zhihan Zhang, Wenhao Yu, Mengxia Yu, Zhichun Guo, and Meng Jiang.
\newblock A survey of multi-task learning in natural language processing: Regarding task relatedness and training methods.
\newblock {\em arXiv preprint arXiv:2204.03508}, 2022.

\bibitem{Ding2023MitigatingTI}
Chuntao Ding, Zhichao Lu, Shangguang Wang, Ran Cheng, and Vishnu~Naresh Boddeti.
\newblock Mitigating task interference in multi-task learning via explicit task routing with non-learnable primitives.
\newblock {\em CVPR}, pages 7756--7765, 2023.

\bibitem{guo2020learning}
Pengsheng Guo, Chen-Yu Lee, and Daniel Ulbricht.
\newblock Learning to branch for multi-task learning.
\newblock In {\em ICML}, pages 3854--3863, 2020.

\bibitem{Zhang2022ATM}
Lijun Zhang, Xiao Liu, and Hui Guan.
\newblock A tree-structured multi-task model recommender.
\newblock In {\em AutoML}, 2022.

\bibitem{Raychaudhuri2022ControllableDM}
Dripta~S. Raychaudhuri, Yumin Suh, Samuel Schulter, Xiang Yu, Masoud Faraki, Amit~K. Roy-Chowdhury, and Manmohan Chandraker.
\newblock Controllable dynamic multi-task architectures.
\newblock {\em CVPR}, pages 10945--10954, 2022.

\bibitem{Yue2023LearningCA}
Zhixiong Yue, Yu~Zhang, and Jie Liang.
\newblock Learning conflict-noticed architecture for multi-task learning.
\newblock In {\em AAAI Conference on Artificial Intelligence}, 2023.

\bibitem{Zhang2019AttentionaugmentedEM}
Zixing Zhang, Bingwen Wu, and Bj{\"o}rn Schuller.
\newblock Attention-augmented end-to-end multi-task learning for emotion prediction from speech.
\newblock {\em ICASSP}, pages 6705--6709, 2019.

\bibitem{huang2022mtl}
Zhiqi Huang, Milind Rao, Anirudh Raju, Zhe Zhang, Bach Bui, and Chul Lee.
\newblock Mtl-slt: multi-task learning for spoken language tasks.
\newblock In {\em Proceedings of the 4th Workshop on NLP for Conversational AI}, pages 120--130, 2022.

\bibitem{wang2023efficient}
Chenguang Wang and Tianshu Yu.
\newblock Efficient training of multi-task neural solver with multi-armed bandits.
\newblock {\em arXiv preprint arXiv:2305.06361}, 2023.

\bibitem{wang2024asp}
Chenguang Wang, Zhouliang Yu, Stephen McAleer, Tianshu Yu, and Yaodong Yang.
\newblock Asp: Learn a universal neural solver!
\newblock {\em IEEE Transactions on Pattern Analysis and Machine Intelligence}, 2024.

\bibitem{kang2011learning}
Zhuoliang Kang, Kristen Grauman, and Fei Sha.
\newblock Learning with whom to share in multi-task feature learning.
\newblock In {\em ICML}, pages 521--528, 2011.

\bibitem{kumar2012learning}
Abhishek Kumar and Hal Daume~III.
\newblock Learning task grouping and overlap in multi-task learning.
\newblock {\em arXiv preprint arXiv:1206.6417}, 2012.

\bibitem{lee2016asymmetric}
Giwoong Lee, Eunho Yang, and Sung Hwang.
\newblock Asymmetric multi-task learning based on task relatedness and loss.
\newblock In {\em International conference on machine learning}, pages 230--238. PMLR, 2016.

\bibitem{lee2018deep}
Hae~Beom Lee, Eunho Yang, and Sung~Ju Hwang.
\newblock Deep asymmetric multi-task feature learning.
\newblock In {\em ICML}, pages 2956--2964, 2018.

\bibitem{zamir2018taskonomy}
Amir~R Zamir, Alexander Sax, William Shen, Leonidas~J Guibas, Jitendra Malik, and Silvio Savarese.
\newblock Taskonomy: Disentangling task transfer learning.
\newblock In {\em CVPR}, pages 3712--3722, 2018.

\bibitem{dwivedi2019representation}
Kshitij Dwivedi and Gemma Roig.
\newblock Representation similarity analysis for efficient task taxonomy \& transfer learning.
\newblock In {\em CVPR}, pages 12387--12396, 2019.

\bibitem{malhotra2022dropped}
Aakarsh Malhotra, Mayank Vatsa, and Richa Singh.
\newblock Dropped scheduled task: Mitigating negative transfer in multi-task learning using dynamic task dropping.
\newblock {\em Transactions on Machine Learning Research}, 2022.

\bibitem{gao2024dmtg}
Yuan Gao, Shuguo Jiang, Moran Li, Jin-Gang Yu, and Gui-Song Xia.
\newblock {DMTG}: One-shot differentiable multi-task grouping.
\newblock In {\em Forty-first International Conference on Machine Learning}, 2024.

\bibitem{li2024Scalable}
Dongyue Li, Aneesh Sharma, and Hongyang~R. Zhang.
\newblock Scalable multitask learning using gradient-based estimation of task affinity.
\newblock In {\em Proceedings of the 30th ACM SIGKDD Conference on Knowledge Discovery and Data Mining}, KDD '24, page 1542–1553, New York, NY, USA, 2024. Association for Computing Machinery.

\bibitem{finn2017model}
Chelsea Finn, Pieter Abbeel, and Sergey Levine.
\newblock Model-agnostic meta-learning for fast adaptation of deep networks.
\newblock In {\em ICML}, pages 1126--1135, 2017.

\bibitem{nichol2018first}
Alex Nichol, Joshua Achiam, and John Schulman.
\newblock On first-order meta-learning algorithms.
\newblock {\em arXiv preprint arXiv:1803.02999}, 2018.

\bibitem{wang2020negative}
Zirui Wang, Zachary~C Lipton, and Yulia Tsvetkov.
\newblock On negative interference in multilingual models: Findings and a meta-learning treatment.
\newblock {\em arXiv preprint arXiv:2010.03017}, 2020.

\bibitem{zhang2019lookahead}
Michael Zhang, James Lucas, Jimmy Ba, and Geoffrey~E Hinton.
\newblock Lookahead optimizer: k steps forward, 1 step back.
\newblock {\em Advances in neural information processing systems}, 32, 2019.

\bibitem{wang2020lookahead}
Jianyu Wang, Vinayak Tantia, Nicolas Ballas, and Michael Rabbat.
\newblock Lookahead converges to stationary points of smooth non-convex functions.
\newblock In {\em ICASSP}, pages 8604--8608, 2020.

\bibitem{zhou2021towards}
Pan Zhou, Hanshu Yan, Xiaotong Yuan, Jiashi Feng, and Shuicheng Yan.
\newblock Towards understanding why lookahead generalizes better than sgd and beyond.
\newblock {\em Advances in Neural Information Processing Systems}, 34:27290--27304, 2021.

\bibitem{byun2022multi}
Ha-Eun Byun, Boeun Kim, and Jay~H Lee.
\newblock Multi-step lookahead bayesian optimization with active learning using reinforcement learning and its application to data-driven batch-to-batch optimization.
\newblock {\em Computers \& Chemical Engineering}, 167:107987, 2022.

\bibitem{mao2021banditmtl}
Yuren Mao, Zekai Wang, Weiwei Liu, Xuemin Lin, and Wenbin Hu.
\newblock Banditmtl: Bandit-based multi-task learning for text classification.
\newblock In {\em Proceedings of the 59th Annual Meeting of the Association for Computational Linguistics and the 11th International Joint Conference on Natural Language Processing (Volume 1: Long Papers)}, pages 5506--5516, 2021.

\bibitem{yu2020gradient}
Tianhe Yu, Saurabh Kumar, Abhishek Gupta, Sergey Levine, Karol Hausman, and Chelsea Finn.
\newblock Gradient surgery for multi-task learning.
\newblock {\em Advances in Neural Information Processing Systems}, 33:5824--5836, 2020.

\bibitem{DBLP:conf/iclr/JavaloyV22}
Adri{\'{a}}n Javaloy and Isabel Valera.
\newblock Rotograd: Gradient homogenization in multitask learning.
\newblock In {\em ICLR}, 2022.

\bibitem{navon2022multi}
Aviv Navon, Aviv Shamsian, Idan Achituve, Haggai Maron, Kenji Kawaguchi, Gal Chechik, and Ethan Fetaya.
\newblock Multi-task learning as a bargaining game.
\newblock {\em arXiv preprint arXiv:2202.01017}, 2022.

\bibitem{kendall2018multi}
Alex Kendall, Yarin Gal, and Roberto Cipolla.
\newblock Multi-task learning using uncertainty to weigh losses for scene geometry and semantics.
\newblock In {\em Proceedings of the IEEE conference on computer vision and pattern recognition}, pages 7482--7491, 2018.

\bibitem{liu2021conflict}
Bo~Liu, Xingchao Liu, Xiaojie Jin, Peter Stone, and Qiang Liu.
\newblock Conflict-averse gradient descent for multi-task learning.
\newblock {\em Advances in Neural Information Processing Systems}, 34:18878--18890, 2021.

\bibitem{liu2021towards}
Liyang Liu, Yi~Li, Zhanghui Kuang, J~Xue, Yimin Chen, Wenming Yang, Qingmin Liao, and Wayne Zhang.
\newblock Towards impartial multi-task learning.
\newblock In {\em ICLR}, 2021.

\bibitem{guangyuan2022recon}
SHI Guangyuan, Qimai Li, Wenlong Zhang, Jiaxin Chen, and Xiao-Ming Wu.
\newblock Recon: Reducing conflicting gradients from the root for multi-task learning.
\newblock In {\em ICLR}, 2022.

\bibitem{DBLP:journals/tmlr/LiuJDJ22}
Shikun Liu, Stephen James, Andrew~J. Davison, and Edward Johns.
\newblock Auto-lambda: Disentangling dynamic task relationships.
\newblock {\em Trans. Mach. Learn. Res.}, 2022, 2022.

\bibitem{gurobi}
{Gurobi Optimization, LLC}.
\newblock {Gurobi Optimizer Reference Manual}, 2023.

\bibitem{kwon2020pomo}
Yeong-Dae Kwon, Jinho Choo, Byoungjip Kim, Iljoo Yoon, Youngjune Gwon, and Seungjai Min.
\newblock Pomo: Policy optimization with multiple optima for reinforcement learning.
\newblock {\em Advances in Neural Information Processing Systems}, 33:21188--21198, 2020.

\bibitem{wu2021autoformer}
Haixu Wu, Jiehui Xu, Jianmin Wang, and Mingsheng Long.
\newblock Autoformer: Decomposition transformers with auto-correlation for long-term series forecasting.
\newblock {\em Advances in Neural Information Processing Systems}, 34:22419--22430, 2021.

\end{thebibliography}

\newpage
\appendix

\section{Proof of Proposition \ref{propos: linear prop}} \label{app: prof pos}
\begin{proof}
Let $\tilde{L}_j(\phi) = L_{j}(\phi, \theta_{j}^{t+1})$ denote the loss function of task $j$ with $\theta_j$ fixed at step $t+1$.
The group transfer gain from $A$ to $j$ is
$$ \mathcal{S}_{A \rightarrow j}^{t} = 1 - \frac{\tilde{L}_j(\phi_{A\cup\{j\}}^{t+1})}{\tilde{L}_j(\phi_{\{j\}}^{t+1})} = \frac{\tilde{L}_j(\phi_{\{j\}}^{t+1}) - \tilde{L}_j(\phi_{A\cup\{j\}}^{t+1})}{\tilde{L}_j(\phi_{\{j\}}^{t+1})}. $$
Similarly, the task transfer gain from $i$ to $j$ is
$$ \mathcal{S}_{i \rightarrow j}^{t} = 1 - \frac{\tilde{L}_j(\phi_{\{i,j\}}^{t+1})}{\tilde{L}_j(\phi_{\{j\}}^{t+1})} = \frac{\tilde{L}_j(\phi_{\{j\}}^{t+1}) - \tilde{L}_j(\phi_{\{i,j\}}^{t+1})}{\tilde{L}_j(\phi_{\{j\}}^{t+1})}. $$

We consider a standard one-step gradient update for the shared parameters $\phi$. Let $G_k^t = \nabla_\phi L_k(\phi^t, \theta_k^t)$ be the gradient of task $k$'s loss with respect to $\phi$ at step $t$. We assume the updates are given by:
\begin{align*} \label{eq:update}
\phi_{\{j\}}^{t+1} &= \phi^t - \eta_t G_j^t \\
\phi_{\{i,j\}}^{t+1} &= \phi^t - \eta_t (G_i^t + G_j^t) \\
\phi_{A\cup\{j\}}^{t+1} &= \phi^t - \eta_t \left(\sum_{k \in A} G_k^t + G_j^t \right)
\end{align*}
This assumes gradients are summed for shared parameters, 
other aggregation schemes are possible but this is common. Task-specific parameters $\theta_k$ are assumed to be updated based only on $L_k$, ensuring $\theta_j^{t+1}$ is the same across different scenarios for evaluating $L_j$.

The differences in shared parameters after one step, relative to $\phi_{\{j\}}^{t+1}$, are:
\begin{equation*}\label{eq:param_diff}
\begin{aligned}
\phi_{\{i,j\}}^{t+1} - \phi_{\{j\}}^{t+1} &= -\eta_t G_i^t \\
\phi_{A\cup\{j\}}^{t+1} - \phi_{\{j\}}^{t+1} &= -\eta_t \sum_{k \in A} G_k^t
\end{aligned}    
\end{equation*}

Assuming $\tilde{L}_j$ is differentiable, we can apply the Mean Value Theorem:
$$ \tilde{L}_j(\phi_{\{j\}}^{t+1}) - \tilde{L}_j(\phi_{X\cup\{j\}}^{t+1}) = \nabla \tilde{L}_j(\xi_X)^T (\phi_{\{j\}}^{t+1} - \phi_{X\cup\{j\}}^{t+1}), $$
where $\xi_X$ is some point on the line segment between $\phi_{\{j\}}^{t+1}$ and $\phi_{X\cup\{j\}}^{t+1}$.
Substituting the parameter differences:
\begin{equation*}
    \begin{aligned} \label{eq:L_diff_i}
\tilde{L}_j(\phi_{\{j\}}^{t+1}) - \tilde{L}_j(\phi_{\{i,j\}}^{t+1}) &= \nabla \tilde{L}_j(\xi_i)^T (-\eta_t G_i^t) = -\eta_t \nabla \tilde{L}_j(\xi_i)^T G_i^t \\
\tilde{L}_j(\phi_{\{j\}}^{t+1}) - \tilde{L}_j(\phi_{A\cup\{j\}}^{t+1}) &= \nabla \tilde{L}_j(\xi_A)^T (-\eta_t \sum_{k \in A} G_k^t) = -\eta_t \nabla \tilde{L}_j(\xi_A)^T \left(\sum_{k \in A} G_k^t\right)
\end{aligned}
\end{equation*}
where $\xi_i$ is between $\phi_{\{j\}}^{t+1}$ and $\phi_{\{i,j\}}^{t+1}$, and $\xi_A$ is between $\phi_{\{j\}}^{t+1}$ and $\phi_{A\cup\{j\}}^{t+1}$.
Now, let's express the transfer gains using these results:
$$ \mathcal{S}_{i \rightarrow j}^{t} = \frac{-\eta_t \nabla \tilde{L}_j(\xi_i)^T G_i^t}{\tilde{L}_j(\phi_{\{j\}}^{t+1})} \quad \text{and} \quad \mathcal{S}_{A \rightarrow j}^{t} = \frac{-\eta_t \nabla \tilde{L}_j(\xi_A)^T \left(\sum_{k \in A} G_k^t\right)}{\tilde{L}_j(\phi_{\{j\}}^{t+1})}. $$

Consider the difference:
\begin{small}
\begin{equation*}
    \begin{aligned}
\mathcal{S}_{A \rightarrow j}^{t} - \frac{1}{|A|}\sum_{i \in A} \mathcal{S}_{i \rightarrow j}^{t} &= \frac{-\eta_t \nabla \tilde{L}_j(\xi_A)^T \left(\sum_{k \in A} G_k^t\right)}{\tilde{L}_j(\phi_{\{j\}}^{t+1})} - \frac{1}{|A|}\sum_{i \in A} \frac{-\eta_t \nabla \tilde{L}_j(\xi_i)^T G_i^t}{\tilde{L}_j(\phi_{\{j\}}^{t+1})} \\
&= \frac{-\eta_t}{\tilde{L}_j(\phi_{\{j\}}^{t+1})} \left[ \nabla \tilde{L}_j(\xi_A)^T \left(\sum_{i \in A} G_i^t\right) - \frac{1}{|A|}\sum_{i \in A} \nabla \tilde{L}_j(\xi_i)^T G_i^t \right] \\
&= \frac{-\eta_t}{\tilde{L}_j(\phi_{\{j\}}^{t+1})} \sum_{i \in A} \left[ \nabla \tilde{L}_j(\xi_A)^T G_i^t - \frac{1}{|A|} \nabla \tilde{L}_j(\xi_i)^T G_i^t \right] \\
&= \frac{-\eta_t}{\tilde{L}_j(\phi_{\{j\}}^{t+1})} \sum_{i \in A} \left( \nabla \tilde{L}_j(\xi_A) - \frac{1}{|A|} \nabla \tilde{L}_j(\xi_i) \right)^T G_i^t.
\end{aligned}
\end{equation*}    
\end{small}

Taking the absolute value and using the assumption $0 < C \le L_j$ for the denominator $\tilde{L}_j(\phi_{\{j\}}^{t+1})$:
$$ \left|\mathcal{S}_{A \rightarrow j}^{t} - \frac{1}{|A|}\sum_{i \in A} \mathcal{S}_{i \rightarrow j}^{t}\right| \le \frac{\eta_t}{C} \left| \sum_{i \in A} \left( \nabla \tilde{L}_j(\xi_A) - \frac{1}{|A|} \nabla \tilde{L}_j(\xi_i) \right)^T G_i^t \right|. $$
By the triangle inequality and Cauchy-Schwarz inequality:
$$ \le \frac{\eta_t}{C} \sum_{i \in A} \left| \left( \nabla \tilde{L}_j(\xi_A) - \frac{1}{|A|} \nabla \tilde{L}_j(\xi_i) \right)^T G_i^t \right| \le \frac{\eta_t}{C} \sum_{i \in A} || \nabla \tilde{L}_j(\xi_A) - \frac{1}{|A|} \nabla \tilde{L}_j(\xi_i) || \cdot ||G_i^t||. $$
Since $L_k$ is $l$-Lipschitz for all $k$, its gradient with respect to $\phi$ is bounded by $l$, i.e., $||\nabla_\phi L_k(\phi, \theta_k)|| \le l$ for all $\phi, \theta_k$. Thus, $||G_i^t|| = ||\nabla_\phi L_i(\phi^t, \theta_i^t)|| \le l$.
Also, $||\nabla \tilde{L}_j(\phi)|| = ||\nabla_\phi L_j(\phi, \theta_j^{t+1})|| \le l$ for any $\phi$.
Using the triangle inequality for the gradient difference:
$$ || \nabla \tilde{L}_j(\xi_A) - \frac{1}{|A|} \nabla \tilde{L}_j(\xi_i) || \le ||\nabla \tilde{L}_j(\xi_A)|| + \frac{1}{|A|} ||\nabla \tilde{L}_j(\xi_i)|| \le l + \frac{1}{|A|} l = l \left(1 + \frac{1}{|A|}\right). $$
Substituting these bounds:
$$ \le \frac{\eta_t}{C} \sum_{i \in A} l \left(1 + \frac{1}{|A|}\right) \cdot l = \frac{\eta_t l^2}{C} \sum_{i \in A} \left(1 + \frac{1}{|A|}\right) = \frac{\eta_t l^2}{C} \left( |A| \cdot 1 + |A| \cdot \frac{1}{|A|} \right) = \frac{\eta_t l^2}{C} (|A| + 1). $$
This completes the proof.
\end{proof}


\section{Formulation Transformation}\label{app: transformation}
We present the transformation of Formulation \ref{eq: formulation} into a Mixed-Integer Quadratic Programming (MIQP) problem with non-linear constraints. This involves introducing a continuous variable $\mathbf{y} \in [0, 1]^n$ and binary variables $Z_{ijk}$ to obtain:
\begin{equation}\label{eq: transformed formulation}
\begin{aligned}
\max_{X,y} \quad & \sum_{j=1}^{m}{\sum_{k=1}^n\sum_{i=1}^n S_{ik} y_jZ_{ijk}} \\
\text{s.t.} \quad 
   &{X^T \mathbf{1}\ge \mathbf{1}}\\ 
   & X \mathbf{1}\ge \mathbf{1} \\
  & (B\odot X)^T\mathbf{1}\le b\\
& (X^T\mathbf{1})\odot\mathbf{y} = \mathbf{1}\\
&Z_{ijk} = X_{ij}\cdot X_{kj}\  \forall i,j,k \\
&||X_{\cdot j_1}-X_{\cdot j_2}||^2\ge 1, j_1\neq j_2 \\
& {X\in\{0,1\}^{n\times m}}
\end{aligned}
\end{equation}
which can be solved using classical solvers; in this work, we employ Gurobi \cite{gurobi}.

\section{Experimental Details}\label{app: detailed result}
This section provides a comprehensive overview of the experimental settings and results, including the datasets descriptions, model architectures used, hyperparameters, benchmark methods, and modifications implemented to ensure a more equitable comparison.

\subsection{Datasets Descriptions}\label{app: datasets}
\textbf{Taskonomy.} Taskonomy is a comprehensive dataset designed to facilitate systematic studies of the relationships among visual tasks. It consists of over 4.5 million images gathered from more than 4,000 indoor scenes. These images are annotated for 26 different tasks including depth prediction, surface normal estimation, and semantic segmentation. Taskonomy aims to support research in multitask learning and transfer learning by providing a wide range of task annotations. It supports five primary vision tasks: Semantic Segmentation (s), Depth Estimation (d), Surface Normal (n), Keypoint Detection (k), and Canny Edge Detection (e). In accordance with the standard criterion utilized in previous studies, these five tasks take precedence in the conduction of experiments.

\textbf{CelebA.} CelebA is a large-scale face attributes dataset, which is widely used for multitask learning involving facial attribute recognition. It contains over 200,000 images of 10,000 celebrities, each annotated with 40 attribute labels such as ``Smiling'', ``Young'', ``Male'', and ``Wearing Hat''. Following the protocol outlined in TAG \cite{fifty2021efficiently}, we select a subset of 9 attributes: 5 o'Clock Shadow, Black Hair, Blond Hair, Brown Hair, Goatee, Mustache, No Beard, Rosy Cheeks, and Wearing Hat, denoted as $\{a_1, a_2, a_3, a_4, a_5, a_6, a_7, a_8, a_9\}$, from the original 40 attributes for experimental purposes.

\textbf{Combinatorial Optimization Benchmarks.} We explore three types of COPs: the Travelling Salesman Problem (TSP), the Capacitated Vehicle Routing Problem (CVRP) and the Orienteering Problem (OP). Two problem scales are considered for each COP: 20 and 50 for TSP, CVRP, and OP. We employ the notation ``COP-scale'', such as TSP-20, to denote a particular task, resulting in a total of 6 tasks.

\textbf{ETTm1.} The ETTm1 dataset comprises detailed recordings from an electricity transformer situated in a specific region of China, spanning data from July 2016 to July 2018. This dataset encompasses six power load series, along with an oil temperature series, recorded at 15-minute intervals. These series are classified as High Useful load (HF), High Useless load (HL), Middle Useful load (MF), Middle Useless load (ML), Low Useful load (LF), and Low Useless load (LL). The dataset is utilized for forecasting purposes, leveraging all series as inputs and focusing on predictions for individual series as separate tasks. Following the settings in MTG \cite{song2022efficient}, it is split into training, validation, and test sets following a 6:2:2 chronological order ratio.

\subsection{Experimental Settings}\label{app: exp setting}
\textbf{Backbones and hyper-parameters.} 
In the computer vision tasks, Taskonomy and CelebA, we employ a ResNet16 encoder coupled with MLP decoders for each task. Model structure hyperparameters and training attributes such as hidden dimensions, encoder layers, initial learning rate, and scheduling method mirror those in \cite{fifty2021efficiently}. For combinatorial optimization benchmarks, we use POMO \cite{kwon2020pomo} as the backbone, with all hyperparameters held constant except for the training episodes, which are set to 10,000. In the domain of time series analysis, the Autoformer architecture is employed as the neural network structure. Time series forecasting encompasses two widely recognized approaches: multivariate and univariate. Given that the ETTm1 dataset comprises seven time series, it is applicable under both frameworks. To facilitate a task grouping experiment, we configure it as several univariate prediction tasks, adapting the Autoformer model to maintain a majority of its components common across tasks while assigning a unique decoder to each task for making predictions. Regarding the detailed hyperparameter settings for the model's structure and training, we adhere to the configurations specified by \cite{song2022efficient}.
 
\textbf{Equipments.} The experiments were conducted on a server equipped with 8 NVIDIA A100 Tensor Core GPU and 128 Intel(R) Xeon(R) Platinum 8358P CPUs. The primary software versions used are CUDA 11.8, TensorFlow 2.14.1, and PyTorch 2.1.2.

\subsection{Further Results on Taskonomy Dataset}\label{app: taskonomy res}

\begin{figure*}[t!]
    \centering         
    \begin{minipage}{\textwidth}
    \centering
    \captionof{table}{Grouing and comparison results on the Taskonomy dataset.} 
\label{tab: results of taskonomy}
\centering
\footnotesize
\setlength{\tabcolsep}{0.35em}
\renewcommand{\arraystretch}{0.8}
\scalebox{.9}{
\begin{tabular}{lc|ccccc|c}
\toprule
& Method     & s & d & n & k & e  & Loss Reduction ($\mathbf{\uparrow}$)\\
\midrule
\midrule
& Naive-MTL & $0.138$ & $0.088$ & $-0.028$ & $-0.077$ & $0.052$ &$ 0.173$ \\
\midrule
\midrule
\multirow{11}{*}{\rotatebox[origin = c]{90}{2 Groups}}
& Random &$-0.033$ &$0.049$ &$0.000$ &$-0.043$ &$0.043$ &$0.016$\\
\cmidrule(lr){2-8}
& \multirow{2}{*}{Optimal} & - & - & $-0.148$ & $0.443$ & - & \multirow{2}{*}{$0.694$}
\\
&  & $0.138$ & $0.088$ & $-0.028$ & $-0.077$ & $0.052$ & 
\\
\cmidrule(lr){2-8}
& \multirow{2}{*}{CS} & $-0.008$ & $-0.690$ & $-0.167$ & - & - & \multirow{2}{*}{$-0.866$} \\
&  & - & - & - & $-0.013$ & $0.012$ &  \\
\cmidrule(lr){2-8}
& \multirow{2}{*}{HOA} & $0.072$ & $0.014$ & $-0.002$ & - & $0.080$ & \multirow{2}{*}{$0.608$} \\
&  & - & - & $-0.148$ & $0.443$ & - &  \\
\cmidrule(lr){2-8}
& \multirow{2}{*}{TAG} & $-0.008$ & $-0.690$ & $-0.167$ & - & - & \multirow{2}{*}{$-0.866$}
\\
&  & - & - & - & $-0.013$ & $0.012$ & 
\\
\cmidrule(lr){2-8}
& \multirow{2}{*}{Ours} & $0.138$ & $0.088$ & $-0.028$ & $-0.077$ & $0.052$ & \multirow{2}{*}{$0.499$}
\\
&  & - & $0.031$ & $-0.068$ & $0.249$ & $0.032$ & 
\\
\midrule
\multirow{16}{*}{\rotatebox[origin = c]{90}{3 Groups}}
& Random &$0.149$ &$0.020$ &$-0.062$ &$0.062$ &$0.020$ &$0.190$\\
\cmidrule(lr){2-8}
& \multirow{3}{*}{Optimal} & $0.199$ & - & $-0.101$ & - & - & \multirow{3}{*}{$0.888$}
\\
&  & - & - & $-0.148$ & $0.443$ & - & 
\\
&  & - & $0.117$ & $0.018$ & - & $0.110$ & 
\\
\cmidrule(lr){2-8}
& \multirow{3}{*}{CS} & $0.052$ & $0.020$ & - & - & - & \multirow{3}{*}{$0.206$} \\
&  & - & $0.146$ & $0.008$ & - & - &  \\
&  & - & - & - & $-0.013$ & $0.012$ &  \\
\cmidrule(lr){2-8}
& \multirow{3}{*}{HOA} & $-0.008$ & $-0.690$ & $-0.167$ & - & - & \multirow{3}{*}{$-0.391$} \\
&  & - & - & $-0.148$ & $0.443$ & - &  \\
&  & - & - & - & $-0.013$ & $0.012$ &  \\
\cmidrule(lr){2-8}
& \multirow{3}{*}{TAG} & $0.052$ & $0.020$ & - & - & - & \multirow{3}{*}{$0.206$}
\\
&  & - & $0.146$ & $0.008$ & - & - & 
\\
&  & - & - & - & $-0.013$ & $0.012$ & 
\\
\cmidrule(lr){2-8}
& \multirow{3}{*}{Ours} & - & - & $-0.101$ & $0.412$ & $0.067$ & \multirow{3}{*}{$0.677$}
\\
&  & - & $0.031$ & $-0.068$ & $0.249$ & $0.032$ & 
\\
&  & $0.138$ & $0.088$ & $-0.028$ & $-0.077$ & $0.052$ & 
\\
\midrule
\multirow{21}{*}{\rotatebox[origin = c]{90}{4 Groups}}
& Random &$0.063$ &$0.031$ &$-0.068$ &$0.249$ &$0.032$ &$0.307$\\
\cmidrule(lr){2-8}
& \multirow{4}{*}{Optimal} & - & - & $0.000$ & - & - & \multirow{4}{*}{$0.833$} \\
&  & $0.199$ & - & $-0.101$ & - & - &  \\
&  & - & $0.146$ & $0.008$ & - & - &  \\
&  & - & - & $-0.101$ & $0.412$ & $0.067$ &  \\
\cmidrule(lr){2-8}
& \multirow{4}{*}{CS} & $0.052$ & $0.020$ & - & - & - & \multirow{4}{*}{$0.206$} \\
&  & - & $0.146$ & $0.008$ & - & - &  \\
&  & $-0.008$ & $-0.690$ & $-0.167$ & - & - &  \\
&  & - & - & - & $-0.013$ & $0.012$ &  \\
\cmidrule(lr){2-8}
& \multirow{4}{*}{HOA} & $0.199$ & - & $-0.101$ & - & - & \multirow{4}{*}{$0.809$} \\
&  & - & $0.146$ & $0.008$ & - & - &  \\
&  & - & - & $-0.148$ & $0.443$ & - &  \\
&  & - & - & - & $-0.013$ & $0.012$ &  \\
\cmidrule(lr){2-8}
& \multirow{4}{*}{TAG}  & $0.052$ & $0.020$ & - & - & - & \multirow{4}{*}{$0.206$} \\
&  & - & $0.146$ & $0.008$ & - & - &  \\
&  & $-0.008$ & $-0.690$ & $-0.167$ & - & - &  \\
&  & - & - & - & $-0.013$ & $0.012$ &  \\
\cmidrule(lr){2-8}
& \multirow{4}{*}{Ours} & - & - & $-0.101$ & $0.412$ & $0.067$ & \multirow{4}{*}{$0.688$} \\
&  & $0.138$ & $0.088$ & $-0.028$ & $-0.077$ & $0.052$ &  \\
&  & - & $0.031$ & $-0.068$ & $0.249$ & $0.032$ &  \\
&  & $0.149$ & - & $-0.062$ & $0.062$ & $0.020$ &  \\
\bottomrule
\end{tabular}
}
    \end{minipage}
\end{figure*}

We compare the proposed methods with Naive-MTL, Random baseline, Optimal baseline, HOA, CS and TAG in Table \ref{tab: results of taskonomy}. The key metric for evaluation is the loss reduction borrowing the results from MTG \cite{song2022efficient}, where a higher value indicates better performance.

From the results in Table \ref{tab: results of taskonomy}, the Naive-MTL method shows a moderate performance with a loss reduction of 0.173.  When splitting 2 groups,  The Random method, CS and TAG perform poorly, resulting in a significant negative loss reduction of -0.866. The HOA method performs the best results except for the optimal baseline with a loss reduction of 0.608. The proposed method (Ours) achieves the second-best performance after the Optimal baseline, with a notable loss reduction of 0.499. 
For the 3 Groups category, the proposed method (Ours) performs strongly, being the best after the Optimal baseline, with a loss reduction of 0.677, while the HOA performs the worst among all baselines with a negative loss reduction. 

Furthermore, the analysis of the table reveals that as the number of groups increases, the performance of our proposed method shows a progressive improvement. This indicates a high level of adaptability and robustness across different task group divisions. Conversely, although the HOA method performs the best in the 2-group case, its performance drops to negative in the 3-group case, demonstrating that it is not as stable across different group numbers.

\subsection{Further Results on CelebA Dataset}\label{app: celeba res}
\begin{figure*}[t!]
    \centering        

    \begin{minipage}{\textwidth}
    \centering
    \captionof{table}{Grouing and comparison results on the CelebA dataset.} 
    \label{tab:results celeba}
    \centering
    \footnotesize
    \setlength{\tabcolsep}{0.35em}
    \renewcommand{\arraystretch}{1}
    \scalebox{.63}{
    \begin{tabular}{lc|ccccccccc|c}
    \toprule
    & Method     & a1 & a2 & a3 & a4 & a5 & a6 & a7 & a8 & a9  & Tot. Error ($\mathbf{\downarrow}$) \\
    \midrule
    \midrule
    & STL                 & $6.47 \pm 0.044$ & $11.27 \pm 0.037$ & $4.19 \pm 0.006$ & $12.29 \pm 0.020$ & $2.72 \pm 0.038$ & $3.12 \pm 0.015$ & $4.98 \pm 0.031$ & $4.85 \pm 0.019$ & $0.73 \pm 0.007$ & $50.621$\\
    \midrule
    \multirow{4}{*}{\rotatebox[origin = c]{90}{MTL}}
    & Naive-MTL           & $6.55 \pm 0.016$ & $11.09 \pm 0.023$ & $4.19 \pm 0.014$ & $12.56 \pm 0.101$ & $2.58 \pm 0.015$ & $3.02 \pm 0.027$ & $4.80 \pm 0.017$ & $4.74 \pm 0.034$ & $0.70 \pm 0.004$ & {$50.233$} \\
    & GradNorm            & $7.18 \pm 0.071$ & $11.35 \pm 0.028$ & $4.21 \pm 0.034$ & $12.18 \pm 0.127$ & $2.52 \pm 0.015$ & $2.85 \pm 0.032$ & $5.01 \pm 0.084$ & $5.29 \pm 0.103$ & $0.72 \pm 0.019$ & $51.312$ \\
    & PCGrad              & $6.58 \pm 0.101$ & $11.12 \pm 0.019$ & $4.27 \pm 0.048$ & $12.67 \pm 0.722$ & $2.61 \pm 0.034$ & $2.92 \pm 0.010$ & $4.96 \pm 0.204$ & $5.02 \pm 0.021$ & $0.69 \pm 0.030$ & $50.835$ \\
    & UW & $6.72 \pm 0.037$ & $11.32 \pm 0.019$ & $4.30 \pm 0.150$ & $13.61 \pm 0.541$ & $2.74 \pm 0.049$ & $2.93 \pm 0.051$ & $5.43 \pm 0.175$ & $4.86 \pm 0.040$ & $0.79 \pm 0.025$ & $52.717$ \\
    \midrule
    \midrule
    \multirow{11}{*}{\rotatebox[origin = c]{90}{2 Groups}}
        & Random            & $6.41 \pm 0.031$ & $11.10 \pm 0.019$ & $4.14 \pm 0.009$ & $12.51 \pm 0.033$ & $2.66 \pm 0.007$ & $2.92 \pm 0.017$ & $4.71 \pm 0.027$ & $4.75 \pm 0.025$ & $0.78 \pm 0.011$ & {$49.987$} \\
    \cmidrule(lr){2-12}
    & \multirow{2}{*}{OPT} & - & $11.16\pm0.002$ & $4.04\pm0.028$ & $12.24\pm0.367$ & - & - & - & - & $0.74\pm0.025$ & \multirow{2}{*}{$49.583$} \\
    &  & $6.60\pm0.017$ & $11.22\pm0.042$ & $4.36\pm0.018$ & $12.05\pm0.133$ & $2.60\pm0.010$ & $2.88\pm0.011$ & $4.82\pm0.037$ & $4.71\pm0.037$ & - &  \\
    \cmidrule(lr){2-12}
    & \multirow{2}{*}{TAG} & $6.50\pm0.158$ & - & - & - & - & - & $4.88\pm0.118$ & - & - & \multirow{2}{*}{$51.036$} \\
    &  & - & $11.16\pm0.053$ & $4.24\pm0.038$ & $13.15\pm0.112$ & $2.66\pm0.004$ & $2.99\pm0.006$ & - & $4.74\pm0.010$ & $0.71\pm0.005$ &  \\
    \cmidrule(lr){2-12}
    & \multirow{2}{*}{CS} & - & - & - & - & $2.59\pm0.026$ & $3.06\pm0.017$ & - & - & - & \multirow{2}{*}{$50.278$} \\
    &  & $6.58\pm0.028$ & $11.15\pm0.027$ & $4.26\pm0.097$ & $12.35\pm0.229$ & $2.54\pm0.029$ & - & $4.86\pm0.059$ & $4.77\pm0.083$ & $0.71\pm0.017$ &  \\
    \cmidrule(lr){2-12}
    & \multirow{2}{*}{HOA} & - & - & - & - & $2.57\pm0.019$ & - & $4.69\pm0.024$ & - & - & \multirow{2}{*}{$49.600$} \\
    &  & $6.55\pm0.018$ & $11.27\pm0.026$ & $4.14\pm0.009$ & $11.87\pm0.045$ & - & $3.03\pm0.016$ & - & $4.81\pm0.025$ & $0.67\pm0.006$ &  \\
    \cmidrule(lr){2-12}
    & \multirow{2}{*}{Ours} & $6.55\pm0.016$ & $11.09\pm0.023$ & $4.19\pm0.014$ & $12.56\pm0.101$ & $2.58\pm0.015$ & $3.02\pm0.027$ & $4.80\pm0.017$ & $4.74\pm0.034$ & $0.70\pm0.004$ & \multirow{2}{*}{$49.534$} \\
    &  & $6.60\pm0.017$ & $11.22\pm0.042$ & $4.36\pm0.018$ & $12.05\pm0.133$ & $2.60\pm0.010$ & $2.88\pm0.011$ & $4.82\pm0.037$ & $4.71\pm0.037$ & - &  \\
    \midrule
    \midrule
    \multirow{16}{*}{\rotatebox[origin = c]{90}{3 Groups}}
    & Random           & $6.65 \pm 0.030$ & $11.16 \pm 0.018$ & $4.12 \pm 0.019$ & $12.16 \pm 0.147$ & $2.61 \pm 0.065$ & $2.90 \pm 0.019$ & $4.78 \pm 0.018$ & $4.71 \pm 0.010$ & $0.79 \pm 0.012$ & $49.886$ \\
    \cmidrule(lr){2-12}
    & \multirow{3}{*}{OPT} & $6.30\pm0.036$ & - & - & - & - & - & $4.67\pm0.013$ & - & $0.71\pm0.009$ & \multirow{3}{*}{$49.323$} \\
    &  & - & $11.16\pm0.002$ & $4.04\pm0.028$ & $12.24\pm0.367$ & - & - & - & - & $0.74\pm0.025$ &  \\
    &  & $6.27\pm0.027$ & - & - & - & $2.59\pm0.006$ & $2.92\pm0.011$ & $4.73\pm0.016$ & $4.72\pm0.019$ & $0.76\pm0.002$ &  \\
    \cmidrule(lr){2-12}
    & \multirow{3}{*}{TAG} & $6.50\pm0.158$ & - & - & - & - & - & $4.88\pm0.118$ & - & - & \multirow{3}{*}{$49.897$} \\
    &  & - & $11.14\pm0.047$ & - & $12.31\pm0.030$ & - & - & - & - & - &  \\
    &  & - & $11.18\pm0.160$ & $4.11\pm0.007$ & - & $2.55\pm0.029$ & $2.92\pm0.063$ & $4.95\pm0.117$ & $4.73\pm0.013$ & $0.76\pm0.014$ &  \\
    \cmidrule(lr){2-12}
    & \multirow{3}{*}{CS} & - & - & - & - & $2.59\pm0.026$ & $3.06\pm0.017$ & - & - & - & \multirow{3}{*}{$50.954$} \\
    &  & $6.50\pm0.158$ & - & - & - & - & - & $4.88\pm0.118$ & - & - &  \\
    &  & - & $11.18\pm0.008$ & $4.10\pm0.013$ & $13.06\pm0.105$ & - & - & - & $4.83\pm0.013$ & $0.74\pm0.004$ &  \\
    \cmidrule(lr){2-12}
    & \multirow{3}{*}{HOA} & - & - & - & - & $2.57\pm0.019$ & - & $4.69\pm0.024$ & - & - & \multirow{3}{*}{$49.716$} \\
    &  & - & $11.14\pm0.047$ & - & $12.31\pm0.030$ & - & - & - & - & - &  \\
    &  & $6.55\pm0.024$ & - & $4.12\pm0.023$ & - & $2.70\pm0.006$ & $2.84\pm0.002$ & $4.89\pm0.021$ & $4.76\pm0.011$ & $0.73\pm0.012$ &  \\
    \cmidrule(lr){2-12}
    & \multirow{3}{*}{Ours} & $6.60\pm0.017$ & $11.22\pm0.042$ & $4.36\pm0.018$ & $12.05\pm0.133$ & $2.60\pm0.010$ & $2.88\pm0.011$ & $4.82\pm0.037$ & $4.71\pm0.037$ & - & \multirow{3}{*}{$49.348$} \\
    &  & $6.55\pm0.016$ & $11.09\pm0.023$ & $4.19\pm0.014$ & $12.56\pm0.101$ & $2.58\pm0.015$ & $3.02\pm0.027$ & $4.80\pm0.017$ & $4.74\pm0.034$ & $0.70\pm0.004$ &  \\
    &  & $6.62\pm0.043$ & $11.32\pm0.131$ & $4.09\pm0.015$ & $12.21\pm0.051$ & $2.60\pm0.044$ & $2.93\pm0.008$ & $4.72\pm0.053$ & - & - &  \\
    \midrule
    \midrule
    \multirow{21}{*}{\rotatebox[origin = c]{90}{4 Groups}}
    & Random  & $6.40 \pm 0.023$ & $11.25 \pm 0.029$ & $4.11 \pm 0.010$ & $12.92 \pm 0.102$ & $2.55 \pm 0.014$ & $3.01 \pm 0.002$ & $4.74 \pm 0.023$ & $4.75 \pm 0.006$ & $0.76 \pm 0.013$ & $50.493$\\
    \cmidrule(lr){2-12}
    & \multirow{4}{*}{OPT} & - & - & $4.08\pm0.079$ & - & - & $2.94\pm0.021$ & $4.82\pm0.033$ & - & $0.74\pm0.006$ & \multirow{4}{*}{$49.169$} \\
    &  & - & $11.29\pm0.129$ & $4.12\pm0.056$ & $12.59\pm0.090$ & - & - & - & - & - &  \\
    &  & $6.30\pm0.036$ & - & - & - & - & - & $4.67\pm0.013$ & - & $0.71\pm0.009$ &  \\
    &  & - & $11.11\pm0.020$ & - & $11.92\pm0.012$ & $2.68\pm0.019$ & - & - & $4.77\pm0.006$ & $0.78\pm0.004$ &  \\
    \cmidrule(lr){2-12}
    & \multirow{4}{*}{TAG} & $6.50\pm0.158$ & - & - & - & - & - & $4.88\pm0.118$ & - & - & \multirow{4}{*}{$49.605$} \\
    &  & - & $11.14\pm0.047$ & - & $12.31\pm0.030$ & - & - & - & - & - &  \\
    &  & - & - & - & - & $2.62\pm0.010$ & $2.99\pm0.004$ & $4.76\pm0.009$ & - & - &  \\
    &  & - & $10.97\pm0.023$ & $4.09\pm0.033$ & $12.40\pm0.054$ & - & - & $4.95\pm0.017$ & $4.63\pm0.012$ & $0.72\pm0.002$ &  \\
    \cmidrule(lr){2-12}
    & \multirow{4}{*}{CS} & - & - & - & - & $2.59\pm0.026$ & $3.06\pm0.017$ & - & - & - & \multirow{4}{*}{$49.753$} \\
    &  & $6.50\pm0.158$ & - & - & - & - & - & $4.88\pm0.118$ & - & - &  \\
    &  & - & $11.14\pm0.047$ & - & $12.31\pm0.030$ & - & - & - & - & - &  \\
    &  & - & $10.97\pm0.023$ & $4.09\pm0.033$ & $12.40\pm0.054$ & - & - & $4.95\pm0.017$ & $4.63\pm0.012$ & $0.72\pm0.002$ &  \\
    \cmidrule(lr){2-12}
    & \multirow{4}{*}{HOA} & - & - & - & - & $2.57\pm0.019$ & - & $4.69\pm0.024$ & - & - & \multirow{4}{*}{$49.853$} \\
    &  & - & $11.14\pm0.047$ & - & $12.31\pm0.030$ & - & - & - & - & - &  \\
    &  & $6.65\pm0.033$ & - & - & - & $2.68\pm0.031$ & - & - & - & - &  \\
    &  & - & $11.18\pm0.160$ & $4.11\pm0.007$ & - & $2.55\pm0.029$ & $2.92\pm0.063$ & $4.95\pm0.117$ & $4.73\pm0.013$ & $0.76\pm0.014$ &  \\
    \cmidrule(lr){2-12}
& \multirow{4}{*}{Ours} & $6.62\pm0.043$ & $11.32\pm0.131$ & $4.09\pm0.015$ & $12.21\pm0.051$ & $2.60\pm0.044$ & $2.93\pm0.008$ & $4.72\pm0.053$ & - & - & \multirow{4}{*}{$49.335$} \\
&  & $6.63\pm0.026$ & $11.36\pm0.083$ & $4.34\pm0.034$ & $12.23\pm0.057$ & $2.77\pm0.019$ & - & $4.94\pm0.009$ & $4.69\pm0.027$ & - &  \\
&  & $6.55\pm0.016$ & $11.09\pm0.023$ & $4.19\pm0.014$ & $12.56\pm0.101$ & $2.58\pm0.015$ & $3.02\pm0.027$ & $4.80\pm0.017$ & $4.74\pm0.034$ & $0.70\pm0.004$ &  \\
&  & $6.60\pm0.017$ & $11.22\pm0.042$ & $4.36\pm0.018$ & $12.05\pm0.133$ & $2.60\pm0.010$ & $2.88\pm0.011$ & $4.82\pm0.037$ & $4.71\pm0.037$ & - &  \\
    \bottomrule
    \end{tabular}
    }
    \end{minipage}
\end{figure*}
We compare our approach against four benchmark MTL methodologies, namely, Naive-MTL, GradNorm, PCGrad, and Uncertainty Weights. Notably, we directly implement the precise groupings presented by \cite{fifty2021efficiently} as the grouping outcome of TAG, HOA, and CS methods, though the total error is based on our testing outcomes. 

As Table \ref{tab:results celeba} exhibits, our method typically surpasses other grouping methodologies and MTL methods, barring OPT groupings, derived from TAG. These may not necessarily represent the optimal groupings in our test results, as our approach outperforms OPT in the 2-split scenario, where our approach achieves a total error of 49.534, with OPT at 49.583. For the 3-split scenario, our approach achieves better results of 49.348 compared to the 2-split case, reaching almost the same performance as the OPT grouping. Furthermore, it continues to exhibit improvement with the 4-split case, achieving a total error of 49.335. As the number of groups increases, TAG and our proposed method demonstrate continuous improvement, while HOA performance declines, and CS exhibits instability.

\subsection{Further Results on Combinatorial Optimization Tasks}\label{app: cop res}
\begin{figure*}[t!]
    \centering        

    \begin{minipage}{\textwidth}
    \centering
    \captionof{table}{Grouing and comparison results on the COP benchmark. } 
\label{tab:results cop}
\centering
\footnotesize
\setlength{\tabcolsep}{0.35em}
\renewcommand{\arraystretch}{1}
\scalebox{1}{
\begin{tabular}{lc|cccccc|c}
\toprule
& Method     & TSP20 & TSP50  & CVRP20 & CVRP50  & OP20 & OP50 & Tot. Gap ($\downarrow$) \\
\midrule
\midrule
&STL &$0.017$\% &$0.277$\% &$0.534$\% &$1.780$\% &$-0.849$\% &$1.117$\% &{$2.876\%$} \\
\midrule
\multirow{6}{*}{\rotatebox[origin = c]{90}{MTL}}
& Naive-MTL & $0.022\%$  & $0.469\%$  & $0.522\%$  & $2.070\%$  & $-0.805\%$  & $1.270\%$  & $3.548\%$  \\
& Bandit-MTL & $0.021\%$ & $0.882\%$ & $0.690\%$ & $2.511\%$ & $-0.865\%$ & $2.114\%$ & $5.354\%$  \\
& PCGrad & $0.028\%$ & $0.708\%$ & $0.605\%$ & $2.411\%$ & $-0.689\%$ & $1.756\%$ & $4.819\%$    \\
& UW & $0.042\%$ & $0.362\%$ & $0.412\%$ & $1.703\%$ & $-0.665\%$ & $1.153\%$ & $3.007\%$  \\
& LS & $0.020\%$ & $0.476\%$ & $0.512\%$ & $2.084\%$ & $-0.792\%$ & $1.197\%$ & $3.498\%$ \\
& Nash-MTL & $0.038\%$ & $0.322\%$ & $0.421\%$ & $1.847\%$ & $-0.873\%$ & $1.279\%$ & $3.034\%$  \\
\midrule
\midrule
\multirow{9}{*}{\rotatebox[origin = c]{90}{3 Groups}}
& Random & $0.031\%$ & $0.535\%$ & $0.416\%$ & $1.741\%$ & $-0.764\%$ & $1.279\%$ & $3.237\%$ \\
\cmidrule(lr){2-9}
& \multirow{3}{*}{OPT} & - & - & $0.404\%$ & $1.692\%$ & - & -  &\multirow{3}{*}{$2.492\%$}   \\
&   & $0.022\%$ & $0.307\%$ & $0.512\%$ & - & - & -    &  \\
&  & - & $0.638\%$ & $0.631\%$ & - & $-0.861\%$ & $0.929\%$  &   \\
\cmidrule(lr){2-9}
& \multirow{3}{*}{TAG} & $0.021\%$ & $0.324\%$ & - & - & - & - & \multirow{3}{*}{$2.807\%$} \\
&  & - & $0.505\%$ & $0.405\%$ & $1.724\%$ & - & - &  \\
&  & - & - & - & - & $-0.771\%$ & $1.104\%$ &  \\
\cmidrule(lr){2-9}
& \multirow{3}{*}{Ours}& $0.021\%$ & $0.324\%$ & - & - & - & -  &\multirow{3}{*}{$2.774\%$}   \\
&  & - & - & $0.404\%$ & $1.692\%$ & - & - &  \\
&  & - & - & - & - & $-0.771\%$ & $1.104\%$ &   \\
\midrule
\midrule
\multirow{14}{*}{\rotatebox[origin = c]{90}{4 Groups}}
& Random & $0.024\%$ & $0.473\%$ & $0.681\%$ & $2.125\%$ & $-0.852\%$ & $1.072\%$ & $3.522\%$ \\
\cmidrule(lr){2-9}
& \multirow{4}{*}{OPT}& - & $0.277\%$ & - & - & - & -  & \multirow{4}{*}{$2.431\%$}  \\
& & - & - & $0.404\%$ & $1.692\%$ & - & -   &\\
&  & $0.027\%$ & $0.415\%$ & - & - & $-0.897\%$ & $1.120\%$   &\\
&  & - & $0.638\%$ & $0.631\%$ & - & $-0.861\%$ & $0.929\%$  &   \\
\cmidrule(lr){2-9}
& \multirow{4}{*}{TAG} & $0.021\%$ & $0.324\%$ & - & - & - & - & \multirow{4}{*}{$2.774\%$} \\
&  & - & $0.458\%$ & $0.512\%$ & - & - & - &  \\
&  & - & - & $0.404\%$ & $1.692\%$ & - & - &  \\
&  & - & - & - & - & $-0.771\%$ & $1.104\%$ &  \\
\cmidrule(lr){2-9}
& \multirow{4}{*}{Ours} & $0.021\%$ & $0.324\%$ & - & - & - & - & \multirow{4}{*}{$2.696\%$}  \\
&  & - & - & $0.404\%$ & $1.692\%$ & - & -  &  \\
&  & $0.034\%$ & - & - & - & $-0.840\%$ & $1.095\%$  &  \\
&  & - & - & - & - & $-0.771\%$ & $1.104\%$   &  \\
\midrule
\midrule
\multirow{16}{*}{\rotatebox[origin = c]{90}{5 Groups}}
& Random & $0.038\%$ & $0.444\%$ & $0.472\%$ & $1.992\%$ & $-0.761\%$ & $1.233\%$ & $3.419\%$\\
\cmidrule(lr){2-9}
& \multirow{5}{*}{OPT} & $0.017\%$ & - & - & - & - & - & \multirow{5}{*}{$2.422\%$} \\
&  & - & $0.277\%$ & - & - & - & - &  \\
&  & - & - & $0.404\%$ & $1.692\%$ & - & - &  \\
&  & $0.027\%$ & $0.415\%$ & - & - & $-0.897\%$ & $1.120\%$ &  \\
&  & - & $0.638\%$ & $0.631\%$ & - & $-0.861\%$ & $0.929\%$ &  \\
\cmidrule(lr){2-9}
& \multirow{5}{*}{TAG} & $0.021\%$ & $0.324\%$ & - & - & - & - & \multirow{5}{*}{$2.757\%$} \\
&  & $0.022\%$ & $0.307\%$ & $0.512\%$ & - & - & - &  \\
&  & - & $0.458\%$ & $0.512\%$ & - & - & - &  \\
&  & - & - & $0.404\%$ & $1.692\%$ & - & - &  \\
&  & - & - & - & - & $-0.771\%$ & $1.104\%$ &  \\
\cmidrule(lr){2-9}
& \multirow{5}{*}{Ours} & $0.021\%$ & $0.324\%$ & - & - & - & - & \multirow{5}{*}{$2.696\%$} \\
&  & - & - & - & - & $-0.771\%$ & $1.104\%$ &  \\
&  & - & - & $0.404\%$ & $1.692\%$ & - & - &  \\
&  & - & - & - & - & - & $1.117\%$ &  \\
&  & $0.034\%$ & - & - & - & $-0.840\%$ & $1.095\%$ &  \\
\bottomrule
\end{tabular}
}
    \end{minipage}
\end{figure*}
In the presented comparative analysis in Table \ref{tab:results cop}, Single-Task Learning (STL) demonstrates a strong baseline with a total gap of 2.876\%, outperforming all Multi-Task Learning (MTL) methods in terms of the Total Gap metric. Specifically, STL exhibits a lower total gap compared to the best-performing MTL method, UW, which has a total gap of 3.007\%. The performance gaps between STL and other MTL methods range from 0.131\% (UW) to 2.478\% (Bandit-MTL), highlighting STL's superior performance in this benchmark.

Within the domain of Task Grouping methods, both TAG and our proposed method demonstrate the ability to surpass the STL baseline in certain aspects. Notably, our method consistently achieves the highest performance among non-optimal baselines across each grouping strategy, indicating its efficacy in handling multiple related tasks simultaneously.

As we consider the performance trends across different task groupings, it is observed that the efficacy of Optimal, TAG, and our method improves as the splits become larger. For instance, in the 3-split task grouping, our method achieves a total gap of 2.774\%, compared to the optimal group's total gap of 2.492\%. This trend suggests that the tasks within the COP benchmark exhibit high positive transfer potential.

In the context of 3-split task grouping, our method achieves logical groupings of tasks, pairing the same types of COPs together: (TSP20, TSP50), (CVRP20, CVRP50), and (OP20, OP50). This reflects an intuitive understanding that tasks of the same types benefit from being trained in concert. Intriguingly, the optimal groupings: (TSP20, TSP50, CVRP20), (CVRP20, CVRP50) and (TSP50, CVRP20, OP20,OP50), 
do not align with these intuitive pairings, suggesting that there may be non-obvious correlations that, when leveraged, could lead to even greater improvements in task performance. 

\subsection{Further Results on Time Series Tasks}\label{app: ts res}

\begin{figure*}[t!]
    \centering        

    \begin{minipage}{\textwidth}
    \centering
    \captionof{table}{Grouing and comparison results on the ETTm1 dataset.} 
\label{tab:results ettm1}
\centering
\footnotesize
\setlength{\tabcolsep}{0.35em}
\renewcommand{\arraystretch}{1}
\scalebox{.8}{
\begin{tabular}{lc|ccccccc|c}
\toprule
& Method     & t1 & t2 & t3 & t4 & t5 & t6 & t7   &MAE ($\mathbf{\downarrow}$)\\
\midrule
\midrule
& STL        & $0.64 \pm 0.016$ & $0.37 \pm 0.009$ & $0.68 \pm 0.015$ & $0.36 \pm 0.011$ & $0.56 \pm 0.027$ & $0.29 \pm 0.002$ & $0.15 \pm 0.006$ & $3.050$ \\
\midrule
\multirow{6}{*}{\rotatebox[origin = c]{90}{MTL}}
& Bandit-MTL & $0.67 \pm 0.041$ & $0.37 \pm 0.010$ & $0.68 \pm 0.051$ & $0.34 \pm 0.003$ & $0.58 \pm 0.017$ & $0.26 \pm 0.004$ & $0.14 \pm 0.003$ & $3.052$ \\
& LS         & $0.67 \pm 0.014$ & $0.38 \pm 0.004$ & $0.67 \pm 0.040$ & $0.36 \pm 0.007$ & $0.58 \pm 0.023$ & $0.28 \pm 0.003$ & $0.15 \pm 0.002$ & $3.094$ \\
& UW         & $0.67 \pm 0.011$ & $0.38 \pm 0.014$ & $0.62 \pm 0.011$ & $0.36 \pm 0.013$ & $0.59 \pm 0.021$ & $0.27 \pm 0.011$ & $0.16 \pm 0.013$ & $3.061$ \\
& Nash-MTL   & $0.61 \pm 0.017$ & $0.38 \pm 0.011$ & $0.65 \pm 0.006$ & $0.36 \pm 0.004$ & $0.59 \pm 0.037$ & $0.27 \pm 0.003$ & $0.15 \pm 0.013$ & $3.004$ \\
& PCGrad     & $0.65 \pm 0.033$ & $0.38 \pm 0.011$ & $0.62 \pm 0.020$ & $0.35 \pm 0.001$ & $0.57 \pm 0.006$ & $0.27 \pm 0.011$ & $0.15 \pm 0.001$ & {$2.997$} \\
& Naive-MTL  & $0.66 \pm 0.014$ & $0.38 \pm 0.006$ & $0.64 \pm 0.056$ & $0.35 \pm 0.009$ & $0.59 \pm 0.016$ & $0.28 \pm 0.005$ & $0.15 \pm 0.009$ & $3.038$ \\
& MTG & $0.69\pm0.028$ & $0.39\pm0.011$ & $0.68\pm0.035$ & $0.36\pm0.015$ & $0.55\pm0.022$ & $0.28\pm0.007$ & $0.15\pm0.000$ & $3.101$ \\
\midrule
\midrule
\multirow{6}{*}{\rotatebox[origin = c]{90}{2 Groups}}
& Random   & $0.66 \pm 0.014$ & $0.38 \pm 0.006$ & $0.64 \pm 0.056$ & $0.35 \pm 0.009$ & $0.59 \pm 0.016$ & $0.28 \pm 0.005$ & $0.15 \pm 0.009$ & $3.035$ \\
\cmidrule(lr){2-10}
& \multirow{2}{*}{OPT} & - & $0.37\pm0.007$ & - & $0.35\pm0.003$ & $0.56\pm0.029$ & - & $0.15\pm0.011$ & \multirow{2}{*}{$2.926$} \\
&  & $0.61\pm0.015$ & - & $0.62\pm0.036$ & $0.36\pm0.007$ & $0.57\pm0.036$ & $0.28\pm0.008$ & - &  \\
\cmidrule(lr){2-10}
& \multirow{2}{*}{TAG} & $0.66\pm0.041$ & $0.37\pm0.005$ & $0.66\pm0.022$ & $0.36\pm0.007$ & - & - & - & \multirow{2}{*}{$3.032$} \\
&  & $0.66\pm0.014$ & $0.38\pm0.006$ & $0.64\pm0.056$ & $0.35\pm0.009$ & $0.59\pm0.016$ & $0.28\pm0.005$ & $0.15\pm0.009$ &  \\
\cmidrule(lr){2-10}
& \multirow{2}{*}{Ours} & - & - & $0.65\pm0.055$ & - & $0.56\pm0.011$ & $0.27\pm0.009$ & $0.16\pm0.006$ & \multirow{2}{*}{$2.989$} \\
&  & $0.61\pm0.015$ & $0.38\pm0.012$ & - & $0.36\pm0.014$ & $0.57\pm0.010$ & - & - &  \\
\midrule
\midrule
\multirow{9}{*}{\rotatebox[origin = c]{90}{3 Groups}}
& Random  & $0.67 \pm 0.025$ & $0.39 \pm 0.004$ & $0.65 \pm 0.031$ & $0.35 \pm 0.004$ & $0.57 \pm 0.020$ & $0.28 \pm 0.009$ & $0.14 \pm 0.001$ & $3.054$ \\
\cmidrule(lr){2-10}
& \multirow{3}{*}{OPT} & - & - & - & $0.35\pm0.018$ & $0.55\pm0.022$ & $0.27\pm0.015$ & - & \multirow{3}{*}{$2.913$} \\
&  & - & $0.37\pm0.007$ & - & $0.35\pm0.003$ & $0.56\pm0.029$ & - & $0.15\pm0.011$ &  \\
&  & $0.61\pm0.015$ & - & $0.62\pm0.036$ & $0.36\pm0.007$ & $0.57\pm0.036$ & $0.28\pm0.008$ & - &  \\
\cmidrule(lr){2-10}
& \multirow{3}{*}{TAG} & $0.64\pm0.016$ & $0.38\pm0.008$ & $0.65\pm0.017$ & - & - & - & - & \multirow{3}{*}{$3.037$} \\
&  & $0.64\pm0.026$ & - & $0.67\pm0.025$ & $0.36\pm0.007$ & $0.58\pm0.021$ & - & - &  \\
&  & $0.63\pm0.030$ & - & $0.66\pm0.037$ & $0.37\pm0.016$ & $0.57\pm0.022$ & $0.29\pm0.016$ & $0.16\pm0.001$ &  \\
\cmidrule(lr){2-10}
& \multirow{3}{*}{Ours} & - & - & $0.65\pm0.055$ & - & $0.56\pm0.011$ & $0.27\pm0.009$ & $0.16\pm0.006$ & \multirow{3}{*}{$2.979$} \\
&  & $0.61\pm0.015$ & $0.38\pm0.012$ & - & $0.36\pm0.014$ & $0.57\pm0.010$ & - & - &  \\
&  & - & $0.37\pm0.007$ & - & - & $0.58\pm0.037$ & $0.27\pm0.012$ & - &  \\
\midrule
\midrule
\multirow{12}{*}{\rotatebox[origin = c]{90}{4 Groups}}
& Random & $0.64 \pm 0.026$ & $0.38 \pm 0.005$ & $0.68 \pm 0.043$ & $0.35 \pm 0.005$ & $0.55 \pm 0.022$ & $0.27 \pm 0.005$ & $0.15 \pm 0.011$ & $3.022$ \\
\cmidrule(lr){2-10}
& \multirow{4}{*}{OPT} & $0.66\pm0.028$ & - & - & - & - & - & $0.14\pm0.005$ & \multirow{4}{*}{$2.906$} \\
&  & - & - & - & $0.35\pm0.018$ & $0.55\pm0.022$ & $0.27\pm0.015$ & - &  \\
&  & - & $0.37\pm0.007$ & - & $0.35\pm0.003$ & $0.56\pm0.029$ & - & $0.15\pm0.011$ &  \\
&  & $0.61\pm0.015$ & - & $0.62\pm0.036$ & $0.36\pm0.007$ & $0.57\pm0.036$ & $0.28\pm0.008$ & - &  \\
\cmidrule(lr){2-10}
& \multirow{4}{*}{TAG} & $0.64\pm0.016$ & $0.38\pm0.008$ & $0.65\pm0.017$ & - & - & - & - & \multirow{4}{*}{$3.026$} \\
&  & $0.63\pm0.004$ & - & $0.63\pm0.030$ & - & $0.58\pm0.025$ & $0.28\pm0.008$ & - &  \\
&  & - & - & $0.68\pm0.039$ & $0.36\pm0.013$ & - & - & - &  \\
&  & - & - & $0.70\pm0.054$ & - & - & - & $0.16\pm0.007$ &  \\
\cmidrule(lr){2-10}
& \multirow{4}{*}{Ours} & - & $0.37\pm0.007$ & - & - & $0.58\pm0.037$ & $0.27\pm0.012$ & - & \multirow{4}{*}{$2.966$} \\
&  & - & - & $0.65\pm0.055$ & - & $0.56\pm0.011$ & $0.27\pm0.009$ & $0.16\pm0.006$ &  \\
&  & - & - & - & - & $0.58\pm0.016$ & $0.27\pm0.008$ & $0.15\pm0.005$ &  \\
&  & $0.61\pm0.015$ & $0.38\pm0.012$ & - & $0.36\pm0.014$ & $0.57\pm0.010$ & - & - &  \\
\bottomrule
\end{tabular}
}
    \end{minipage}
\end{figure*}

Table \ref{tab:results ettm1} reveals that our approach outperforms other task grouping strategies across all divisions. Among the multitask learning (MTL) methods evaluated, PCGrad exhibits the best performance with an MAE of 2.997, outperforming other strategies such as Bandit-MTL (3.052) and Naive-MTL (3.038). However, these MTL methods still yield higher MAE values compared to our approach.

Three task grouping strategies were examined: Random, OPT, and TAG. The Random strategy achieves an MAE of 3.035, comparable to Naive-MTL. OPT exhibits the best performance in the 2-split scenario, with an MAE of 2.926, while TAG performs comparably to Random, with an MAE of 3.032.

The proposed method consistently outperforms all other methods with the exception of the Optimal across various task split scenarios. Specifically, in the 4-split scenario, the proposed method achieves the lowest overall MAE of 2.966, demonstrating its robustness and effectiveness in handling complex multitask learning applications. Notably, the MAE improvements over the best-performing MTL method (PCGrad) and the Random grouping strategy are 0.031 and 0.056, respectively.

\subsection{Ablation Study on Transfer Gain and Task Grouping Solver}
    In this section, we conduct an ablation study to highlight the significance of the novel transfer gain and the mathematical programming framework, compared to TAG \cite{fifty2021efficiently}. To demonstrate the importance of our transfer gain metric, we apply the mathematical framework from Formulation \ref{eq: formulation} for task grouping using $\mathcal{S}_{i \rightarrow j}^{t}$ as per Formulation \ref{def: ta}, and TAG's affinity $\mathcal{Z}_{i \rightarrow j}^{t}$. This approach is termed ``Ours-MP'' for our method and ``TAG-MP'' for TAG's method. We also evaluate our framework's performance against TAG's branch and bound techniques, introducing ``Ours-BB'' for the Branch \& Bound method guided by the transfer gain $\mathcal{S}_{i \rightarrow j}^{t}$.
  
\begin{figure}[t!]
\centerline{\includegraphics[width=\columnwidth]{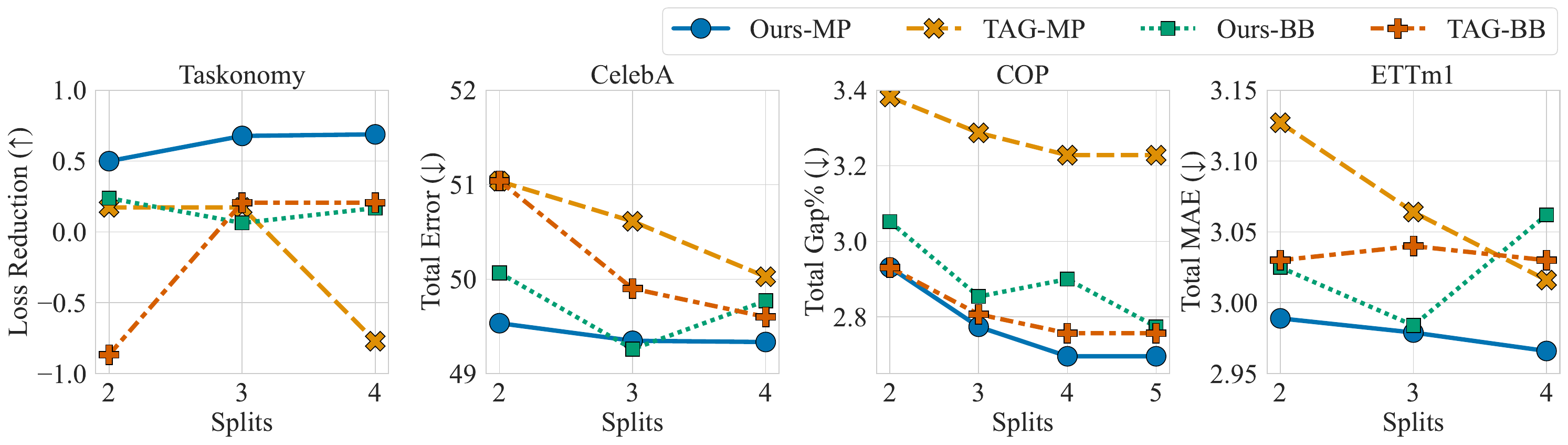}}
    \caption{Performance comparison of Ours-MP, TAG-MP, Ours-BB and TAG-BB across multiple grouping splits on Taskonomy, CelebA, COP, and ETIm1 benchmarks. 
    }
    \label{fig:compare tag}
\end{figure}

    \textbf{Ablation on Transfer Gain.}
    The first aspect of the comparison focuses on the transfer gain between $\mathcal{S}_{i \rightarrow j}^{t}$ and $\mathcal{Z}_{i \rightarrow j}^{t}$, as implemented in Ours-MP and TAG-MP, respectively. Results in Figure \ref{fig:compare tag} reveal that Ours-MP consistently surpasses TAG-MP across all benchmarks. This is evidenced by its greater loss reduction in the Taskonomy dataset, lower total error in the CelebA dataset, minimized Total Gap percentage in COP, and reduced Total MAE in ETIm1 with an increasing number of grouping splits. TAG-MP's declining performance, particularly in the CelebA dataset, suggests that the transfer gain proposed in our methodology more accurately captures the task relationships than the one proposed in TAG, under a consistent task grouping solver.

    \textbf{Ablation on Task Grouping Solver.}
    In Figure \ref{fig:compare tag}, Ours-MP demonstrates better performance than Ours-BB, providing results across all grouping splits. 
    
    \begin{wraptable}{r}{7cm}
    \vspace{-4mm}
      \caption{Comparative Analysis of Time Efficiency Across Different Task Splits and Dataset. ``s'', ``m'' and ``h'' stand for seconds, minutes and hours, respectively. ``$\times$'' indicates that the method fails to solve the problem within an 8-hour time limit.}
      \label{compare time}
      \centering
          \setlength{\tabcolsep}{0.35em}
        \renewcommand{\arraystretch}{1.5}
        \scalebox{0.65}{
      \begin{tabular}{c|cc|cc|cc|cc}
        \toprule
         & \multicolumn{2}{c|}{Taskonomy (5)}  &\multicolumn{2}{c|}{CelebA (9)}  & \multicolumn{2}{c|}{COP (6)} & \multicolumn{2}{c}{ETTm1 (7)} \\
         \midrule
        Splits & BB & MP & BB & MP & BB & MP & BB & MP  \\
        \midrule
        2 &0.002s  &0.110s &0.312s  &0.907s  &0.007s &0.085s  &0.024s  &0.122s   \\
        3 &0.046s  &0.563s &1.93m  &2.255s  &0.279s &0.713s  &2.005s  &1.078s   \\
        4 & 0.431s &1.253s & $\times$ &4.009s  &5.955s &0.899s  &1.63m  &1.677s   \\
        5 & 2.454s & 1.733s & $\times$ &7.261s  &1.48m &1.621s  & 1.02h &2.218s   \\
        6 & - & - & $\times$ &21.44s  &16.96m &5.013s  & $\times$ &4.158s   \\
        \bottomrule
        \end{tabular}}
      \vspace{-4mm}
    \end{wraptable}
    We further demonstrate the detailed time-cost for Ours-MP and Ours-BB in Table \ref{compare time}, labeled as MP and BB, respectively. The results indicate that MP exhibits better time efficiency and scalability across different task splits and datasets compared to BB. Specifically, MP generally scales more effectively with an increasing number of splits than the baseline method BB. In particular, for CelebA, our method demonstrates exceptional coverage, capable of handling all scenarios up to six splits efficiently within 30 seconds. When examining the COP dataset, it is evident that the time cost for the BB method increases substantially, potentially exponentially, with larger splits, or fails to deliver results within an eight-hour limit. This trend is also observable across datasets; within the same split categories of two and three, the time cost for BB grows drastically with the increase in task numbers. This highlights our method's scalability and robustness in managing increased computational demands across varying scenarios.

\subsection{Visualized Transfer Gains}
\begin{figure}[ht]
\centering
\subcaptionbox{Taskonomy\label{fig:tky heatmap}}
[.35\linewidth]{\includegraphics[width=\linewidth]{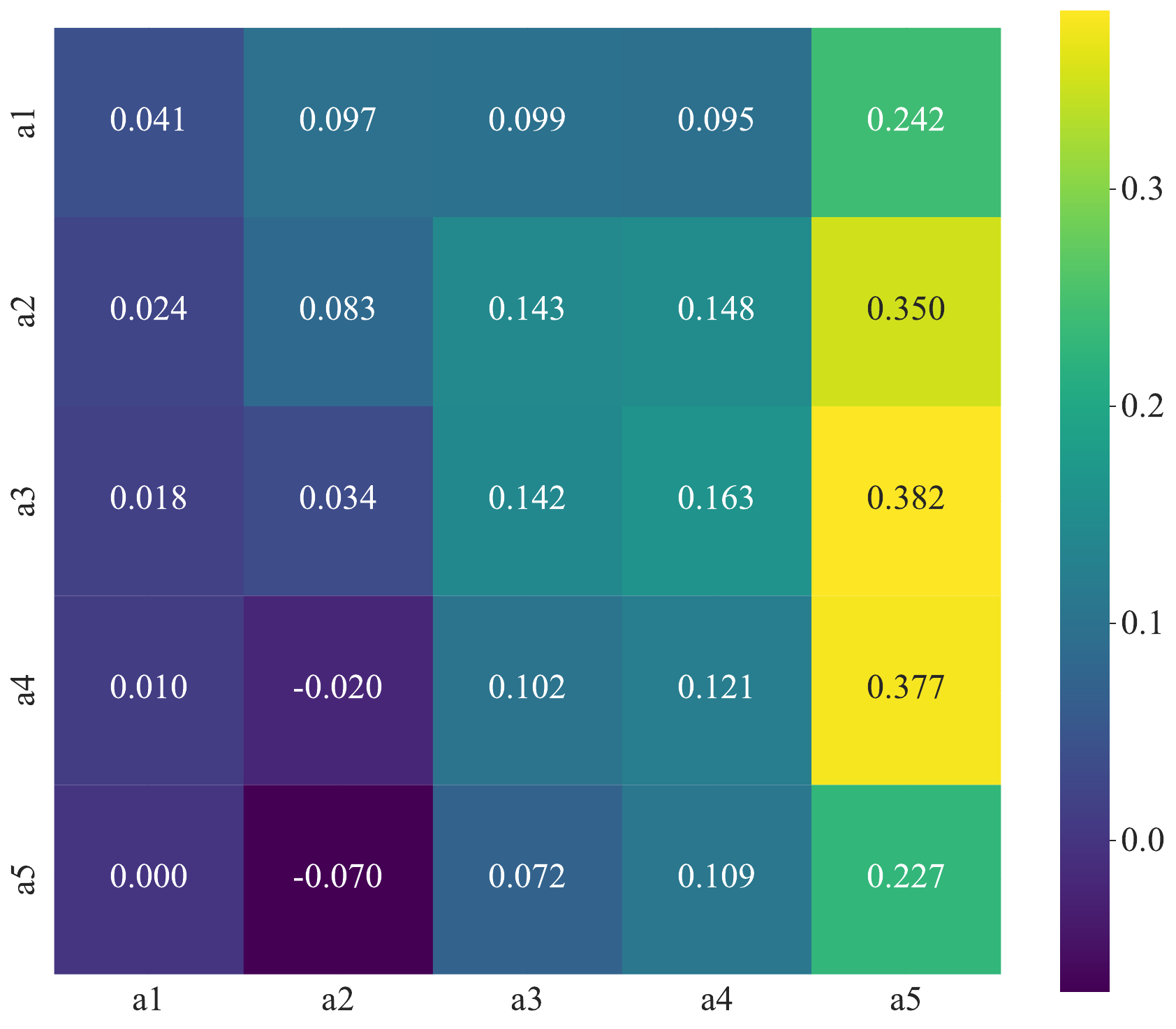}}
\subcaptionbox{CelebA\label{fig:celeba heatmap}}
[.35\linewidth]{\includegraphics[width=\linewidth]{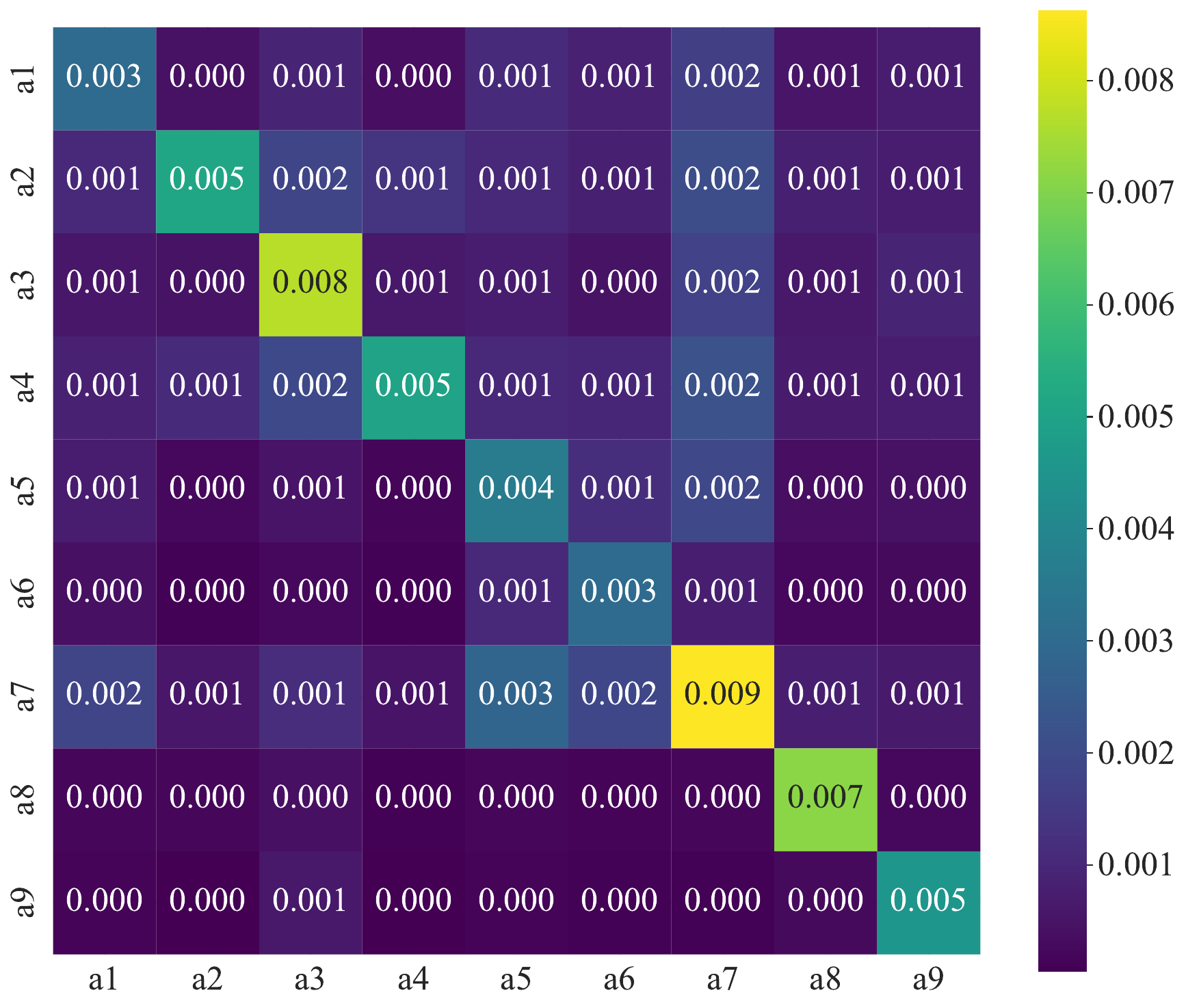}} \\
\subcaptionbox{COP\label{fig:cop heatmap}}
[.35\linewidth]{\includegraphics[width=\linewidth]{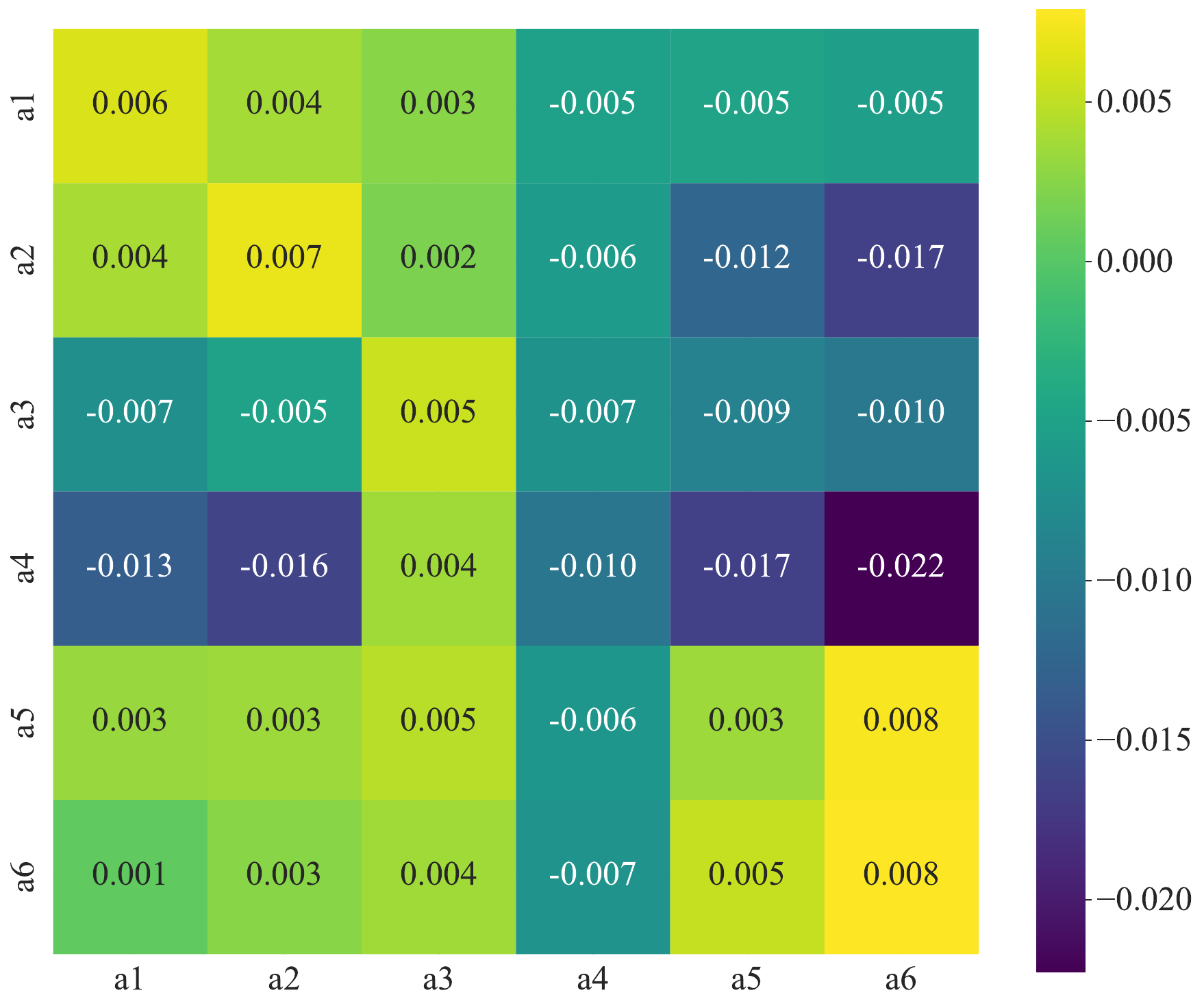}}
\subcaptionbox{ETTm1\label{fig:ett heatmap}}
[.35\linewidth]{\includegraphics[width=\linewidth]{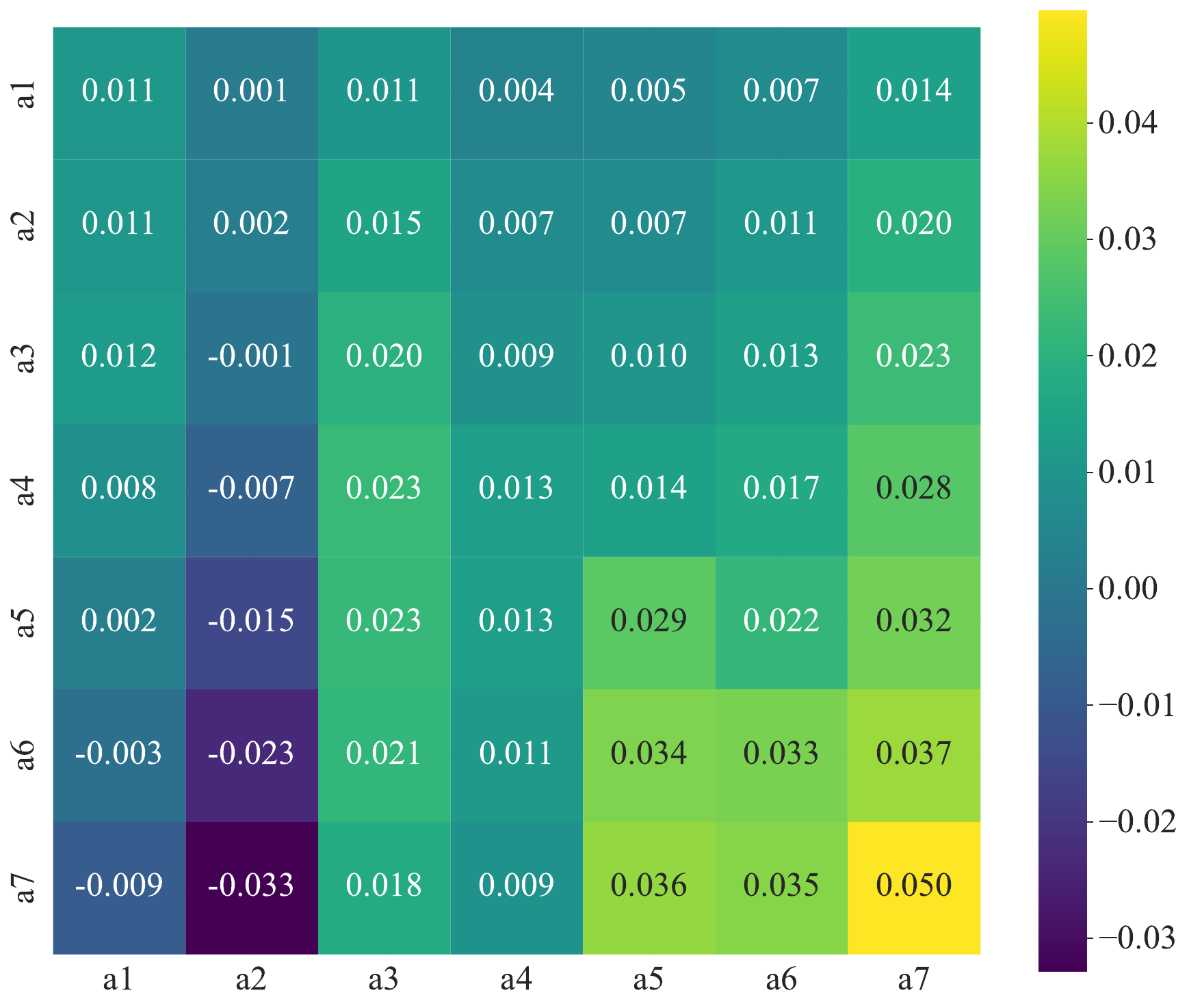}}
\caption{Demonstration of the pairwise transfer gains collect by our method, where each row represents the gain from the corresponding task to other tasks.}
\label{fig:heatmaps}
\end{figure}
The pairwise transfer gains collected by our groups are visualized in Figure~\ref{fig:heatmaps}. The heatmap reveals that nearly all tasks achieve the highest transfer gain with themselves, while exhibiting varying transfer gain distributions across other tasks. This variation underscores an opportunity for improving the overall performance through a task grouping method. Moreover, certain subsets within the heatmap demonstrate overall high transfer gains among tasks, suggesting these tasks form an effective group naturally. It is also noteworthy that the heatmap for ETTm1 (Figure~\ref{fig:ett heatmap}) is less structured compared to those for the other two tasks. This lack of structure introduces additional challenges in generating effective grouping results, thereby presenting more difficult scenarios for task grouping.

\section{Discussions on Computational Complexity}\label{app: comput. complexity}
In this section, we provide detailed derivations and additional experimental results related to the computational complexity of the proposed method. 

\subsection{Derivations}\label{app: derivation}

\begin{figure*}[t!]
    \centering        
    \begin{minipage}{\textwidth}
    \centering
    \captionof{table}{Computation complexity of basic operators.} 
\label{tab:complexity}
\setlength{\tabcolsep}{0.35em}
\centering
\renewcommand{\arraystretch}{1}
\scalebox{1}{
\begin{tabular}{cccccc}
\toprule
 & Task Num
 & \multicolumn{1}{c}{Avg. Dim. Param.} & \multicolumn{1}{c}{Avg. Complexity of FF} & \multicolumn{1}{c}{Avg. Complexity of BP}
\\
\midrule
& $n$ & $\mathcal{C}$  & $\mathcal{F}$  & $\mathcal{B}$  \\
\bottomrule
\end{tabular}
}
    \end{minipage}
    \vspace{3mm}
    \begin{minipage}{\textwidth}
    \centering
    \captionof{table}{Complexity Computation} 
\label{tab:complexity comput.}
\setlength{\tabcolsep}{0.35em}
\centering
\renewcommand{\arraystretch}{1}
\scalebox{.78}{
\begin{tabular}{rccccccc}
\toprule
&Method & Loss Comput.
 & \multicolumn{1}{c}{Grad. Comput.} & \multicolumn{1}{c}{Update Params.} & \multicolumn{1}{c}{High-order Loss} & In Total
\\
\midrule
&TAG & $n\mathcal{F}$ & $n\mathcal{B}$  & $n\mathcal{C}$  & $n^2\mathcal{F}$  & $(n^2+n)\mathcal{F}+n\mathcal{B}+n\mathcal{C}$\\
&Ours & $n\mathcal{F}$ & $n\mathcal{B}$  & $\frac{n(n+3)}{2}\mathcal{C}$  & $(n^2+n)\mathcal{F}$  & $(n^2+2n)\mathcal{F}+n\mathcal{B}+\frac{n(n+3)}{2}\mathcal{C}$\\
&Ours (Sampling) & $n\mathcal{F}$ & $n\mathcal{B}$  & $\frac{(n+1)(n+5)}{6}\mathcal{C}$  & $\frac{(n+1)(n+2)}{3}\mathcal{F}$  & $\frac{n^2+6n+2}{3}\mathcal{F}+n\mathcal{B}+\frac{(n+1)(n+5)}{6}\mathcal{C}$\\
\bottomrule
\end{tabular}
}
    \end{minipage}
\end{figure*}

Following the notations in Table \ref{tab:complexity}, we derive the computational complexity as follows: There are four parts of computation to collect the transfer gains for TAG and our method: {(1)} Loss computation $\{L_{j}(\phi_{\{j\}}^{t}, \theta_{j}^{t}),\forall j\}$ needs the computation of $n\mathcal{F}$ for both methods; 
{(2)} Gradient Computation, $\{\nabla L_{j}(\phi_{\{j\}}^{t}, \theta_{j}^{t}),\forall j\}$ needs the computation of $n\mathcal{B}$ for both methods; 
{(3)} Update parameters: $\{(\phi_{\{j\}}^{t+1}, \theta_{j}^{t}),\forall j\}$ for TAG and $\{(\phi_{\{i,j\}}^{t+1}, \theta_{j}^{t+1}), (\phi_{\{j\}}^{t+1}, \theta_{j}^{t+1}),\forall i, j\}$ for our method, resulting the computation of $n\mathcal{C}$ and $(\frac{(1+n)n}{2}+n)\mathcal{C}$, respectively;
{(4)} High-order loss computation: $\{L_{j}(\phi_{\{i\}}^{t+1}, \theta_{j}^{t}),\forall i, j\}$ for TAG and $\{L_{j}(\phi_{\{j\}}^{t+1}, \theta_{j}^{t}), L_{j}(\phi_{\{i,j\}}^{t+1}, \theta_{j}^{t+1}), \forall i,j\}$ for our method, resulting the computation of $n^2\mathcal{F}$ and $(n+n^2)\mathcal{F}$, respectively.
Detailed comparison of computation results are demonstrated in Table \ref{tab:complexity comput.}.

\subsection{Complexity of Sampling Strategy}\label{app:sampling strategy}
Based on the results in Appendix \ref{app: derivation}, although the computational complexity of our approach is on the same order of magnitude as TAG, the time required for transfer gains collection may become a computational bottleneck when the number of tasks is large. To address this issue, we propose a sampling-based method: Define a random variable $T$ that follows a uniform distribution and denote as $T \sim \text{Unif}(\{1, 2, ..., n\})$. During the training, we randomly select a subset of tasks with the size of $T$ and transfer gains are gathered solely from this subset. Then the the computational cost of collecting transfer gains is:
$$
\mathbb{E}\left[\frac{T(T+3)}{2}\mathcal{C} + (T^2+T)\mathcal{F}\right] = \frac{(n+1)(n+5)}{6}\mathcal{C}+\frac{(n+1)(n+2)}{3}\mathcal{F},
$$
which significantly reduces the computational cost of our method and is substantially lower than that of TAG because $\mathcal{C}\ll\mathcal{F}$ in practice. 

\subsection{Further Results of Lazy Collection Strategy}\label{app: impact of freq}
\begin{figure}[ht]
\centering
\subcaptionbox{CelebA\label{fig:celeba trend}}
[0.48\linewidth]{\includegraphics[width=\linewidth]{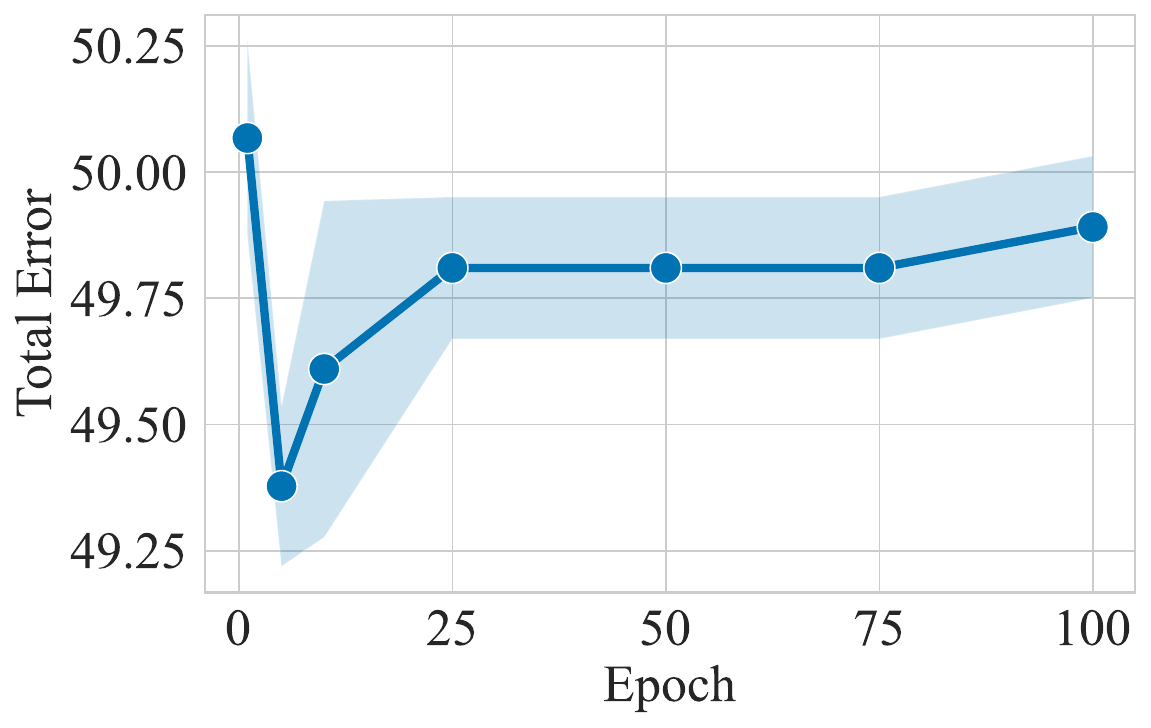}}
\subcaptionbox{ETTm1\label{fig:cop trend}}
[0.48\linewidth]{\includegraphics[width=\linewidth]{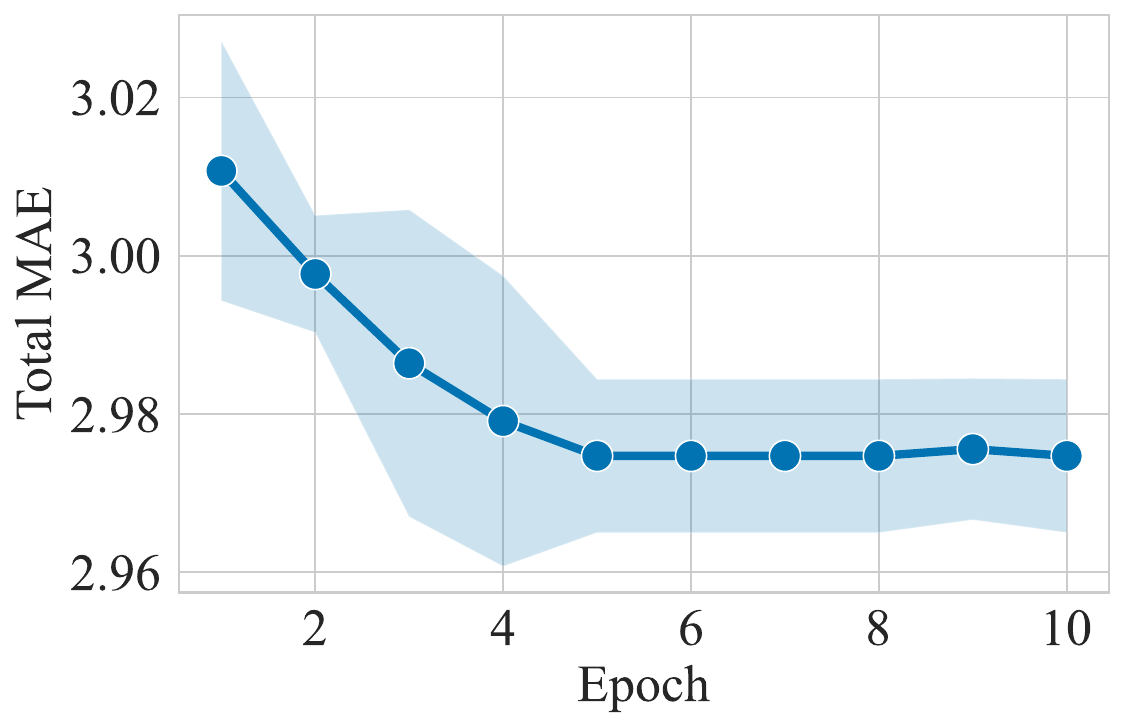}}
\caption{\label{fig:partial}This graph illustrates the performance trends of groups generated by our method, as a function of the number of epochs involved in the construction of transfer gains. Each point represents the mean total error at a specific epoch.}
\end{figure}

In this part, we investigate the impact of the number of epochs required for our proposed method on the final performance metrics.

Figure \ref{fig:partial} delineates the performance trends across different epochs for the CelebA and ETTm1 datasets. For CelebA, the total error exhibits a marked decrease as the number of epochs increases, stabilizing after approximately 20 epochs. This stabilization suggests that the model quickly benefits from initial training iterations but reaches a plateau, indicating little to no gain from additional training beyond this point. The shaded area.
Conversely, the ETTm1 dataset shows a more gradual decline in total Mean Absolute Error (MAE) as the number of epochs grows. The initial drop in MAE is quite steep, suggesting significant learning gains from early training. Subsequently, the MAE curve flattens out after about 6 epochs, which implies that further training yields diminishing improvements in model performance.
These findings demonstrate that the accumulation of transfer gains does not occur uniformly throughout all training periods; instead, it varies, with certain training phases yielding more substantial enhancements than others.

Combining the analysis in Section \ref{sec: efficiency impr.} and above results, we can deduce several empirical guidelines for optimizing the collection of transfer gains with regard to efficiency: \textbf{(1)} The initial training period is crucial for uncovering task relationships, indicating the importance of concentrating resources on the early stages of training; \textbf{(2)} A range of 5-50 steps is considered optimal for gathering transfer gains, as it is probable that the gains from consecutive steps will be similar.

\section{Broader Impact}\label{app: broader imp.}
This research on Multi-Task Learning (MTL) presents a novel approach to task grouping that achieves significant efficiency gains in both academic and industrial settings. It stands out for its flexibility in adapting to diverse and realistic demands, which is crucial for managing complex tasks efficiently. This adaptability is particularly important in the context of growing computational demands in large-scale data analysis.
However, the approach also brings forth ethical considerations. The interpretation of inter-task affinities, if not handled cautiously, could lead to incorrect associations or biases, especially in sensitive contexts. It is imperative to recognize and address these risks to prevent potential misuse and ensure the responsible application of this technology.
Despite these considerations, the method's ability to considerably reduce computational demands, while catering to specific requirements and maintaining high accuracy, is a noteworthy advancement in MTL.

\end{document}